\PassOptionsToPackage{numbers}{natbib} 
\documentclass{article} % For LaTeX2e
\usepackage{paper_style,times}

\usepackage[utf8]{inputenc} % allow utf-8 input
\usepackage[T1]{fontenc}    % use 8-bit T1 fonts
\usepackage{url}            % simple URL typesetting
\usepackage{booktabs}       % professional-quality tables
\usepackage{amsfonts}       % blackboard math symbols
\usepackage{nicefrac}       % compact symbols for 1/2, etc.
\usepackage{microtype}      % microtypography
\usepackage{xcolor}         % colors
\usepackage{marvosym}
\usepackage{fontawesome}
\usepackage{pifont}
\usepackage{multirow}
\usepackage{graphicx}
\usepackage{amsmath}
\usepackage{amssymb}
\usepackage{algorithm,algpseudocode}
\usepackage{wrapfig}
\usepackage{booktabs}
\usepackage{xcolor}
\usepackage{titlesec}  % 提供 \titleformat
\usepackage{titletoc}  % 提供 \startcontents / \printcontents / \titlecontents
\usepackage[most]{tcolorbox}
\usepackage{xcolor}
\usepackage{enumitem}
\usepackage{tikz} % 仅用于标题左侧的小圆点
\setcitestyle{square,numbers,comma,sort&compress}
\usepackage{enumitem}
\usepackage{colortbl} % 提供 \rowcolor
\usepackage[table]{xcolor} % 定义灰色等颜色
\usepackage{tabularx} % 导言区

\definecolor{myviolet}{HTML}{9B8AFB}
\definecolor{grayblue}{RGB}{170, 180, 230}
\definecolor{lightblue}{RGB}{150, 200, 255}
\definecolor{citecolor}{HTML}{0071bc}
\usepackage[pagebackref=false,breaklinks=true,letterpaper=true,colorlinks,citecolor=citecolor,bookmarks=false]{hyperref}
\usepackage{titlesec}
\titlespacing*{\subsection}{0pt}{0.1\baselineskip}{0.25\baselineskip}
\titlespacing*{\section}{0pt}{0.1\baselineskip}{0.1\baselineskip}
\titlespacing*{\subsection}{0pt}{0.1\baselineskip}{0.15\baselineskip}

\titlespacing{\section}{0pt}{*0.1}{*0.1}
\titlespacing{\subsection}{0pt}{*0.1}{*0.1}
\titlespacing{\subsubsection}{0pt}{*0.1}{*0.1}
\usepackage{caption}
\newcommand{\cmark}{\ding{51}} % ✓
\newcommand{\xmark}{\ding{55}} % ✗
\newcommand{\ie}{\textit{i.e.,}} % ✓
\newcommand{\eg}{\textit{e.g.,}} % ✓

\newcommand{\tablesize}{\fontsize{8pt}{8pt}\selectfont}

\newcommand{\our}{Paper2Video}
\newcommand{\bench}{Paper2Video}
\newcommand{\agent}{PaperTalker}
\title{\includegraphics[width=0.7cm, height=0.7cm]{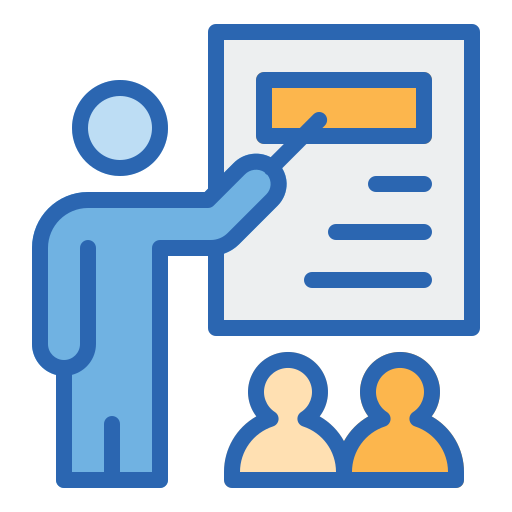} Paper2Video:
Automatic Video Generation from Scientific Papers}

\author{Zeyu Zhu\textsuperscript{*}\quad
Kevin Qinghong Lin\textsuperscript{*}\quad 
Mike Zheng Shou\textsuperscript{\Letter}\\
\\
Show Lab, National University of Singapore \\
}

\iclrfinalcopy
\begin{document}

\maketitle
\let\thefootnote\relax
\footnotetext{$^*$ Equal contribution, $^{\textrm{\Letter}}$ Corresponding author.}
% \let\thefootnote\relax
% \footnotetext{$^{\textrm{\Letter}}$ Corresponding author.}
\vspace{-2\baselineskip} 
\begin{abstract}
\vspace{-1\baselineskip} 
Academic presentation videos have become an essential medium for research communication, yet producing them remains highly labor-intensive, often requiring hours of slide design, recording, and editing for a short 2 to 10 minutes video. 
Unlike natural video, presentation video generation involves distinctive challenges: long-context inputs from research papers, dense multi-modal information (text, figures, tables), and the need to coordinate multiple aligned channels such as slides, subtitles, speech, and human talker. 
To address these challenges, we introduce \textbf{\bench}, the first benchmark of 101 research papers paired with author-created presentation videos, slides, and speaker metadata. We further design four tailored evaluation metrics—Meta Similarity, PresentArena, \textit{PresentQuiz}, and \textit{IP Memory}—to measure how videos convey the paper's information to the audience and affect the work impact. Building on this foundation, we propose \textbf{\agent}, the first multi-agent framework for academic presentation video generation. It integrates slide generation with effective layout refinement by a novel effective \textit{Tree Search Visual Choice}, cursor grounding, subtitling, speech synthesis, and talking-head rendering, while parallelizing slide-wise generation for efficiency. Experiments on \bench~demonstrate that the presentation videos produced by our approach are more faithful and informative than existing baselines, establishing a practical step toward automated and ready-to-use academic video generation. Our dataset, agent, and code are available at \url{https://github.com/showlab/Paper2Video}.

\end{abstract}

% Introduction
% story line:
% multi-modal long-document conditioned video generation
% motivation:
% manually create video for research present is labor-intensive
% much demand for explanatory videos to amplify research impact
% challenge
% no benchmark and metric
% ?

% [teaser]
% 1. reflect present video (paper input + slide + sub + talking head + cursor)
% 2. 
\vspace{-1\baselineskip} 
\section{Introduction}
\vspace{-0.3\baselineskip} 
% [impact background] improtance of academic presentation video creation, how larbor-intensive it is. --> automatic presentation video generation is improtant and practially usefull.
% [compared with existing method] agent for sub-task, natural video gen
% [what we did] benchmark, p2vagent, evalution

% ##[impact background] automatic  presentation video is labor-intensive important for academic communication, but it is labor-intensive including slide creation, talking head per slide / recording, subtitling;
% average spend a hour to develop a 5-10mins. 
% ##[compared with existing method] Despite there has some work for auto PPT, poster, coding but automatically presentation video generation being a promising direction and less exploration.
% compare with natural video gen, presentation video is unique because
% 1. sourced from long-context paper (interleaved, multi-page, professional), information compression;
% 2. it pair with multiple aligned channel / multi-tasks: slide creation , text2speech, subtitling, talking head, cursor. project-level
% 3. lack of evaluation and metric. what define a good presentation video, considering the knowledge conveying via visual format, audio-friendly; 
% In this work, we focus on this problem, particualry, we address two core problems:
% 1. how to automatically create a presentation video from paper;
% for this 
% Paper2Video
% A:
% B:

% 2. how to evaluate a presentation video.
% for this, 
% PaperTalk-101
Academic presentation videos are widely used in research communication, serving as a crucial and effective means to bridge researchers, as many conferences require them as an essential material for submission. However, the manual creation of such a video is highly labor-intensive, requiring slide design, subtitle writing, per-slide recording, and careful editing, which on average may take several hours to produce a $2$ to $10$ minute video for a scientific paper. Despite some prior works on slide and poster generation~\cite{sun2021d2s,zheng2025pptagent,pang2025paper2poster} and other AI4Research tasks~\cite{ai4research,writing_ass,goldsack2022making,paper2code,paper2agent}, automatic academic presentation video generation is a superproblem of them, a practical yet more challenging direction. 

\begin{figure}[!t]  % h=here, t=top, b=bottom, p=单独浮动页，!=放宽限制
    \vspace{-1\baselineskip}
    \centering
    \small
    \captionsetup{skip=2pt}  
    \includegraphics[width=\linewidth]{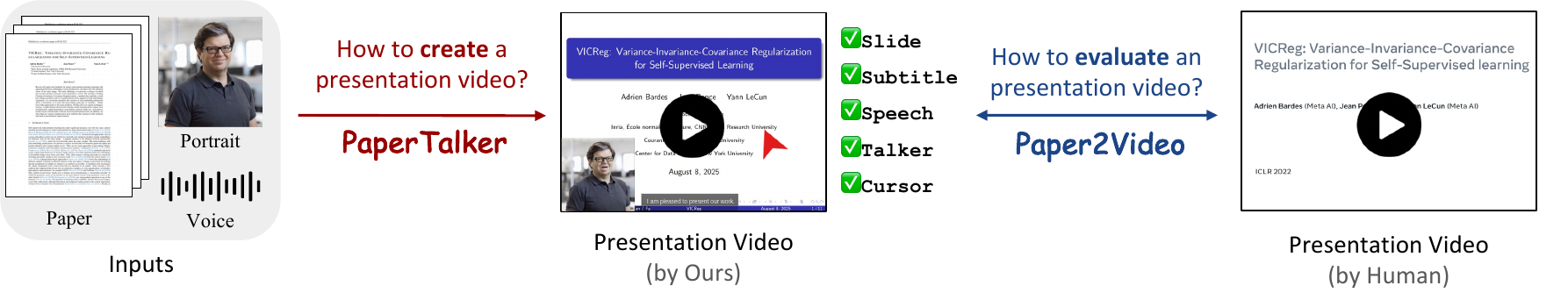}
    \caption{{This work solves two core problems for academic presentations:} \textbf{Left:} \textit{how to create a presentation video from a paper?} {\agent} -- an agent integrates slide, subtitling, cursor grounding, speech synthesis, and talking-head video rendering. \textbf{Right:} \textit{how to evaluate a presentation video?} {\bench} -- a benchmark with well-designed metrics to evaluate presentation quality.}
    \label{fig:teaser}
    \vspace{-0.5\baselineskip}
\end{figure}

 % such as Veo3~\cite{deepmind2025veo3}, 
 \vspace{-0.2\baselineskip} 
Unlike natural video generation~\cite{sd_video,show_1,wan,deepmind2025veo3}, presentation video exhibits distinctive characteristics, including multi-sensory integration, multi-figure conditioning, and high text density, which highlight the limitations of current natural video generation models~\cite{ma2025controllable}. Specifically, academic presentation video generation faces several crucial challenges:
\textit{a.} It originates from long-context papers that contain dense text as well as multiple figures and tables;
\textit{b.} It requires the coordination of multiple aligned channels, including slide generation~\cite{zheng2025pptagent}, subtitling, text-to-speech~\cite{tts-f5}, cursor control, and talking head generation~\cite{fantasytalking,cui2024hallo2};
\textit{c.} It lacks well-defined evaluation metrics: what constitutes a good presentation video, particularly in terms of knowledge conveyance and audience accessibility. Even for the state-of-the-art end-to-end video–audio generation model Veo3~\cite{deepmind2025veo3}, notable limitations remain in video length, clarity of dense on-screen text, and multi-modal long-document condition.
In this work, we try to solve these two core problems as shown in Figure~\ref{fig:teaser}.
%{\textbf{(i)} What defines a good academic presentation video}, and {\textbf{(ii)} How to effectively automatically generate such a video from a research paper}.

\vspace{-0.2\baselineskip} 
To enable comprehensive evaluation of academic presentation video generation, we present the \textbf{\bench} Benchmark, comprising $101$ paired research papers and author-recorded presentation videos from recent conferences, together with original slides and speaker identity metadata. Based on this benchmark, we develop a suite of metrics to comprehensively evaluate generation quality from multiple dimensions: 
\textbf{(i)} Meta Similarity — We employ a VLM to evaluate the alignment of generated slides and subtitles with human-designed counterparts.
\textbf{(ii)} PresentArena — We use a VideoLLM as a proxy audience to perform double-order pairwise comparisons between generated and human-made videos.
Notably, the primary purpose of a presentation is to \textit{effectively convey the information contained in the paper}. To this end, we introduce \textbf{(iii) PresentQuiz}, which treats the VideoLLMs as the audience and requires them to answer paper-derived questions given the videos. Furthermore, another important purpose of presentation video is to \textit{enhance the visibility and impact of the author’s work}. Motivated by real-conference interactions, we introduce \textbf{(iv) IP Memory}, which measures how well an audience can associate authors and works after watching presentation videos.

% \textbf{(iii) Academic IP} — Inspired by real conference scenarios, this metric examines whether, after watching a video, researchers can recall the work and formulate relevant questions, thereby measuring how memorable and academically impactful the presentation video is.

% \kevin{here, quite long, need to have (i) (ii) (iii) to structure the process;}
\vspace{-0.2\baselineskip} 
To effectively generate ready-to-use academic presentation videos, we propose \textbf{\agent}, the first multi-agent framework that enables academic presentation video generation from research papers and speaker identity. It integrates subsequent key modules: \textbf{(i)} Slide Generation. Instead of adopting the commonly used format (\textit{e.g.}, pptx, XML) from a template slide as in ~\cite{zheng2025pptagent}, we employ LaTeX code for slide generation from sketch, given its formal suitability for academic use and higher efficiency. Specifically, we employ a state-of-the-art Coder to generate code and introduce an effective \textbf{focused debugging} strategy, which iteratively narrows the scope and resolves compilation errors using feedback that indicates the relevant rows.
%\kevin{effective \textbf{debugging} strategy}. 
To address the insensitivity of LLMs to fine-grained numerical adjustments, we propose a novel method called \textbf{Tree Search Visual Choice}. This approach systematically explores parameter variations to generate multiple branches, which are then concatenated into a single figure. A VLM is then tasked with selecting the optimal branch, thereby effectively improving element layouts such as figure and font size.
\textbf{(ii)} Subtitling and Cursor Grounding. We generate subtitles and cursor prompts for each sentence based on the slides. Then we achieve cursor spatial-temporal alignment using \textbf{Computer-use grounding model}~\cite{lin2025showui, qin2025ui} models and WhisperX~\cite{bain2023whisperx} respectively. \textbf{(iii)} Speech Synthesis and Talking-head Rendering. We synthesize personalized speech via text-to-speech models~\cite{chen2024f5} and produce talking-head videos~\cite{cui2024hallo2,fantasytalking} for author presentations. Inspired by human recording practice and the independence between each slide, we \textbf{parallelize generation} across slides, achieving a speedup of more than $\mathbf{6\times}$.
% \kevin{comment camel} Our multi-agent framework is implemented within the \texttt{CAMEL}\footnote{https://github.com/camel-ai/camel},  promoting simplicity and enabling scalability. 
We will open-source all our data and codebase to empower the research community.

\vspace{-0.4\baselineskip} 
To summarize, our contributions are as follows:
\vspace{-0.6\baselineskip} 
\begin{itemize}[leftmargin=*]
    \item We present \bench, the first high-quality benchmark of $101$ papers with author-recorded presentation videos, slides, and speaker metadata, together with evaluation metrics: Meta Similarity, PresentArena, PresentQuiz, and IP Memory.
    
    \vspace{-0.2\baselineskip} 
    \item We propose {\agent}, the first multi-agent framework for academic presentation video generation. It introduces three key modules: \textbf{(i)} tree search visual choice for fine-grained slide generation; \textbf{(ii)} a GUI-grounding model coupled with WhisperX for spatial-temporal aligned cursor grounding; and \textbf{(iii)} slide-wise parallel generation to improve efficiency.

    % \kevin{include a/b/c/d}, key insights such as visual mcts; leverage cua / whisperx for spatial-temporal-grounded alignment; to speedup, parallel 
    % the first multi-agent framework for academic presentation video generation, featuring unified multi-channel integration, CoT-based slide debugging with visual inspection, and parallel video synthesis for efficiency.

    \vspace{-0.2\baselineskip} 
    \item Results on {\bench} confirm the effectiveness of {\agent}, which outperforms human-made presentations by 10\% in PresentQuiz accuracy and achieves comparable ratings in user studies, indicating that its quality approaches that of human-created content.
    
    % which achieves \textbf{more than 10\% higher} PresentArena score, PresentQuiz accuracy, and IP Memory than top-performing baselines such as Veo3. [comparable with human-made]
    % PresentArena score & human study - [comparable with human-made] 

    % \kevin{add details number regarding our optimization and improvement here\kevin{sota video-audio gen like veo3 demonstrate significant limitation (xx\%). meanwhile, our agent beat veo3, how sim. to human, other multi-agent }}
\end{itemize}

% to summary, our contributions are:
% 1. Benchmark
% % 1
% 2. For effective, agent core design -- module A: debug from code, enable it effectively debugging with visual inspection
% 3. For effecient, we design a parallel generation -- module B, which speedup the heavy talking head generation with \%x, 

% % 2
% 2. we propose Paper2Video agent, it is the first multi-agent workflow, which integrate XXX tools work together, enable automatically without human;
% 3. To enhance coding-rich visual task, we develop a better debugging strategy,

% 4. Expes demonstratate, metric, 
% Video Creation Agent - MovieAgent, PresentAgent
% Visual Design Agent - SlideGen, PosterGen
\vspace{-0.5\baselineskip} 
\section{Related Works}
\vspace{-0.3\baselineskip} 
% \kevin{size / vid. dur / domain / diff + code}
% 需 booktabs, multirow 宏包；如需更紧凑标题距离，可用 \usepackage{caption}
\begin{table}[!t]
\centering
\begingroup
\setlength{\tabcolsep}{4pt}       % 列间距（默认约 6pt）
\renewcommand{\arraystretch}{0.95}% 行距（默认 1.0）
% \captionsetup{skip=2pt}         % 若已加载 caption 宏包，可启用以缩小标题间距
\footnotesize
\caption{\textbf{Comparison of \our~with existing benchmarks.} Top: existing natural video generation; Button: recent Agents for research works.
% where they support subtitle, slides and cursor generation, and presenter identity image and audio control.
}
\footnotesize
\begin{tabular}{@{}lccccccc@{}} % 去掉左右内边距；移除列格式中的空格
\toprule
\multirow{2}{*}{Benchmarks} & 
\multirow{2}{*}{Inputs} & 
\multirow{2}{*}{Outputs} & 
\multirow{2}{*}{Subtitle} & 
\multirow{2}{*}{Slides} & 
\multirow{2}{*}{Cursor} & 
\multicolumn{2}{c}{Speaker} \\
\cmidrule(lr){7-8}
 & & & & & & Face & Voice \\
\midrule
\multicolumn{8}{c}{{\textcolor[RGB]{105, 105, 105}{\textit{Natural Video Generation}}}}  \\
VBench~\cite{vbench} & Text & Short Vid. & \xmark & \xmark & \xmark & \xmark & \xmark \\
VBench$+$$+$~\cite{vbench++} & Text\&Image & Short Vid. & \xmark & \xmark & \xmark & \xmark & \xmark\\
Talkinghead~\cite{talking-head-1} & Audio\&Image & Short Vid. & \xmark & \xmark & \xmark & \textcolor{citecolor}{\cmark} & \textcolor{citecolor}{\cmark} \\
MovieBench~\cite{wu2025moviebench} & Text\&Audio\&Image & Long Vid. & \textcolor{citecolor}{\cmark} & \xmark & \xmark & \textcolor{citecolor}{\cmark} & \textcolor{citecolor}{\cmark} \\
\midrule
\multicolumn{8}{c}{{\textcolor[RGB]{105, 105, 105}{\textit{Multimodal Agent for Research}}}}  \\
% Paperbench~\cite{paperbench} & Paper & Code & \xmark & \xmark & \xmark & \xmark & \xmark \\
% Paper2Code~\cite{paper2code} & Paper & Code & \xmark & \xmark & \xmark & \xmark & \xmark \\
Paper2Poster~\cite{pang2025paper2poster} & Paper & Poster & \xmark & \xmark & \xmark & \xmark & \xmark \\
PPTAgent~\cite{zheng2025pptagent} & Doc.\&Template & Slide & \xmark & \textcolor{citecolor}{\cmark} & \xmark & \xmark & \xmark \\
PresentAgent~\cite{shi2025presentagent} & Doc.\&Template & Audio\&Long Vid. & \textcolor{citecolor}{\cmark} & \textcolor{citecolor}{\cmark} & \xmark & \xmark & \xmark \\
\bench~(Ours) & Paper\&Image\&Audio & Audio\&Long Vid. & \textcolor{citecolor}{\cmark} & \textcolor{citecolor}{\cmark} & \textcolor{citecolor}{\cmark} & \textcolor{citecolor}{\cmark} & \textcolor{citecolor}{\cmark} \\
\bottomrule
\end{tabular}
\endgroup
\end{table}

\subsection{Video Generation}
\vspace{-0.2\baselineskip} 
% video gen development
% 1. natural short video -- <10 sec; diffusion model
% 2. natural long video -- movieagent, agent colloration + diffusion model
% 3. professional acamide video -- agent colloration + \textbf{code gen.} as academic video require accurate / serious layout, control such as distance, timestamp, so we found code gen (usually bug-free)
Recent advances in video diffusion models~\cite{sd_video,wan,vbench,vbench++} have substantially improved \textit{natural} video generation in terms of length, quality, and controllability. However, these \textbf{end-to-end} diffusion models still struggle to produce long videos~\cite{deepmind2025veo3,fantasytalking} (\textit{e.g.}, several minutes), handle multiple shots, and support conditioning on multiple images~\cite{ma2025controllable}. 
Moreover, most existing approaches generate only video without aligned audio, leaving a gap for real-world applications. To address these limitations, recent works leverage \textbf{multi-agent} collaboration to generate multi-shot, long video–audio pairs and enable multi-image conditioning. 
Specifically, for natural videos, MovieAgent~\cite{wu2025automated} adopts a hierarchical CoT planning strategy and leverages LLMs to simulate the roles of a director, screenwriter, storyboard artist, and location manager, thereby enabling long-form movie generation. Alternatively, PresentAgent~\cite{shi2025presentagent} targets presentation video generation but merely combines PPTAgent~\cite{zheng2025pptagent} with text-to-speech to produce narrated slides. 
% \textit{academic videos}
However, it lacks personalization (\textit{e.g.}, mechanical speech and absence of a presenter) and fails to generate academic-style slides (\textit{e.g.}, missing opening and outline slides), thereby limiting its applicability in academic contexts. Our work addresses these limitations and enables ready-to-use academic presentation video generation.

% \kevin{Our work}

% \subsection{Agent for Research}
\vspace{-0.5\baselineskip} 
\subsection{AI for Research}
\vspace{-0.8\baselineskip} 
% https://arxiv.org/abs/2507.01903
% \kevin{more richer}
Many useful tasks have been explored under the umbrella of AI for Research (AI4Research)~\cite{ai4research}, which aims to support the full scholarly workflow spanning text~\cite{dasigi-etal}, static visuals~\cite{pang2025paper2poster}, and dynamic video~\cite{shi2025presentagent}. With the breakthrough of LLMs in text generation and the Internet search ability, extensive efforts have been devoted to academic writing~\cite{writing_ass} and literature surveying~\cite{sci_lit,deyoung2021ms2,multi-xscience,goldsack2022making}, substantially improving research efficiency. Besides, some works~\cite{paperbench,scireplicate} benchmark AI agents’ end-to-end ability to replicate top-performing ML papers, while others leverage agents to enable idea proposal~\cite{Llm-srbench} and data-driven scientific inspiration~\cite{scienceagentbench,bixbench}. 
To further enhance productivity, a growing number of work focuses on the automatic visual design of figures~\cite{IconShop}, slides~\cite{zheng2025pptagent}, posters~\cite{pang2025paper2poster}, and charts~\cite{hu2024novachart}. More recently, Paper2Agent~\cite{paper2agent} has reimagined research papers as interactive and reliable AI agents, designed to assist readers in understanding scientific works. However, very few studies have investigated video generation for scientific purposes, leaving this area relatively underexplored. Our work belongs to one of the pioneering efforts in this direction, initiating a systematic study on academic presentation video generation.

% 1. LLM:
% agent for writing assistance, agent for survey

% 2. Visual static, image
% slide gen, poster gen, chart bar gen

% 3. Visual dynamic, video
% auto. instructional video, our work belong a pironner work in this category.

% Benchmark
% task formulation 
% data curation
% metrics
% Perception – [which is better] ✅
% Similarity – [video, slides+subtitle, speech] ✅
% Performance - [video QA] ✅
% Ablation experiments
% talking-head&personalized speech – [needle in a haystack] ✅
% 4 video clips + 4 qa & querry image or audio
% cursor (?)
% user study on w/o cursor, w/o talking-head, w/o personalized speech
% 导言区：
% \usepackage{graphicx}
% \usepackage{subcaption}

% 导言区：\usepackage{graphicx}\usepackage{subcaption}

% \section{\our~Benchmark}
% \begin{figure}[t]
%     \centering
%     \includegraphics[width=\linewidth]{figure/data_stat.pdf}
%     \caption{Overview of Paer2Video benchmark.\kevin{all pdf; by three separate fig;}}
%     \label{fig:data_stat}
% \end{figure}
\vspace{-0.5\baselineskip} 
\section{\bench~Benchmark}
\vspace{-0.5\baselineskip} 
\subsection{Task Definition}
\vspace{-0.5\baselineskip} 
% \kevin{slightly longer}
Given a research paper and the author’s identity information, our goal is to automatically synthesize an academic presentation video that faithfully conveys the paper’s core contributions in an audience-friendly manner. 
We identify that a perfect presentation video is usually required to integrate four coordinated components: 
\textbf{(\textit{i}) slides} contain well-organized, visually oriented, expressive figures and tables with concise text description;
%\kevin{well-organized content, visually rather than dense text}
\textbf{(\textit{ii}) synchronized subtitles and speech} are semantically aligned with the slides, including supplementary details;
% \kevin{is usually not exactly the same as slide}
\textbf{(\textit{iii}) presenter} should exhibit natural yet professional facial expressions, ideally accompanied by appropriate gestures;
% \kevin{encourage to show face, sometimes gesture}
and \textbf{(\textit{iv}) a cursor indicator} serves as an attentional anchor, helping the audience focus and follow the narration.
%kevin{that anchors visual attention, helps the audience to focus}. 

\vspace{-0.2\baselineskip}
This task poses several distinctive challenges: 
\textbf{\textit{a}. Multi-modal Long-Context Understanding.} Research papers span many pages with dense text, equations, figures, and tables.
% Generating formal, well-structured slides with fine-grained layout requires content selection, cross-modal grounding, and fine-grained layout design. 
\textbf{\textit{b}. Multi-turn Agent Tasks.} It is challenging to solve this task with a single end-to-end model, as it requires multi-channel generation and alignment (\textit{e.g.}, slides, cursors, and presenter). 
%An efficient, well-designed agent system is therefore needed to address this task.
\textbf{\textit{c}. Personalized Presenter Synthesis.} 
%The human presenter shapes credibility and audience engagement. However,
Achieving high-quality, identity-preserving, and lip-synchronous talking-head video remains time-consuming, and even more challenging when jointly modeling voice, face, and gesture.
\textbf{\textit{d}. Spatial-Temporal-Grounding.} 
%The cursor is a crucial visual cue for the audience to follow the narrative. However, 
Producing cursor trajectories synchronized with narration and slide content demands precise alignment between linguistic units and visual anchors. 

% what define a good present video, require / challenges:
% 0. paper long ctx inputs
% 1. slide gen (visual layout)
% 2. text2speech ()
% 3. talking head (slow, consider mouth, hand gesture, formal)
% 4. cursor 
\vspace{-0.5\baselineskip} 
\subsection{Data Curation}
% non-trivial
% diversity, topics, year, paired human video 
% how we combine to obtain a full video
% human filtering

% topic cv \%, nlp \%;
% percentage of with slide /without slide
\vspace{-0.5\baselineskip}
\textbf{Data Source.} We use AI conference papers as the data source for two reasons: (i) they offer high-quality, diverse content across subfields with rich text, figures, and tables; and (ii) the field’s rapid growth and open-sharing culture provide plentiful, polished author-recorded presentations and slides on YouTube and SlidesLive. However, complete metadata are often unavailable (\textit{e.g.}, presentation videos, slides, presenter images, and voice samples). We thus manually select papers with relatively complete metadata and supplement missing fields by sourcing presenter images from authors’ websites. Overall, we curate 101 peer-reviewed conference papers from the past three years: 41 from machine learning (\textit{e.g.}, NeurIPS, ICLR, ICML), 40 from computer vision (\textit{e.g.}, CVPR, ICCV, ECCV), and 20 from natural language processing(\textit{e.g.}, ACL, EMNLP, NAACL).
Each instance includes the paper’s full \LaTeX{} project and a matched, author-recorded presentation video comprising the slide and talking-head streams with speaker identity (\textit{e.g.}, portrait and voice sample). For 40\% of the data, we additionally collect the original slide files (PDF), enabling direct, reference-based evaluation of slide generation.

% \kevin{But most of paper does not pair with complete metadata (); Therefore, we manually xxx}
\vspace{-0.2\baselineskip}
\textbf{Data Statistics.} 
Overall, {\bench} covers 101 paper-video pairs spanning diverse topics as shown in Figure~\ref{fig:stat} (a), ensuring broad coverage across fields. The paper contains $13.3K$ words($3.3K$ tokens), 44.7 figures, and 28.7 pages on average, serving as multi-modal long document inputs. As illustrated in Figure~\ref{fig:stat} (b) and (c), we also report the distributions of slides per presentation and video durations in {\bench}.  On average, presentations contain 16 slides and last 6min~15s, with some samples reaching up to 14 minutes. Although {\bench} comprises 101 curated presentations, the benchmark is designed to evaluate long-horizon agentic tasks rather than mere video generation.

% \kevin{refer paper2poster stats.}
% \kevin{highlight human effort;}
\newcounter{sqctr}
\newcommand{\sqimg}[2]{%
  \stepcounter{sqctr}%
  \begin{minipage}[t]{0.31\textwidth}
    \centering
    \includegraphics[width=\linewidth,height=\linewidth,keepaspectratio]{#1}
    \par\vspace{2pt}\small(\alph{sqctr})~#2%
  \end{minipage}%
}
\begin{figure}[t]
  \centering
  \setcounter{sqctr}{0} % 每次图开始时清零

  \sqimg{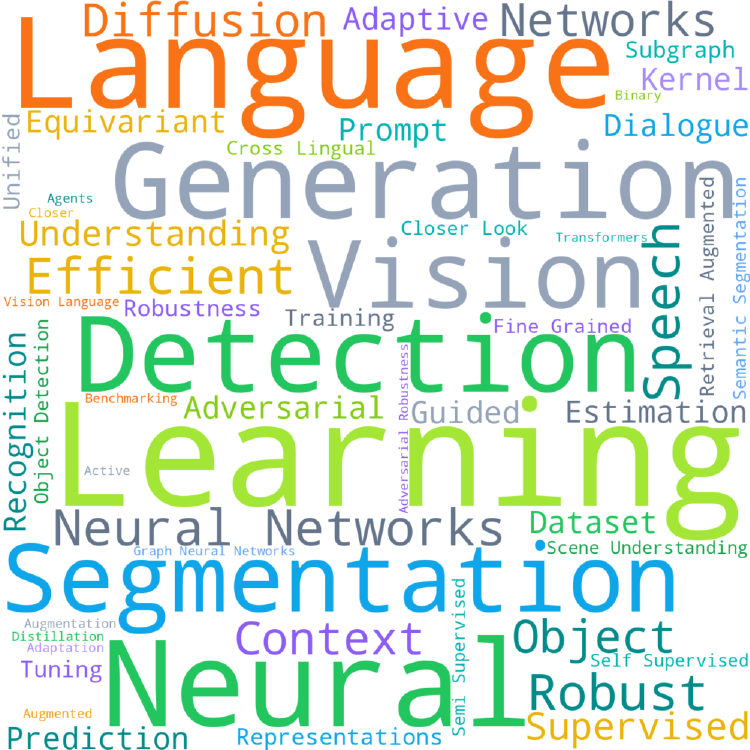}{Word cloud of topics}\hfill
  \sqimg{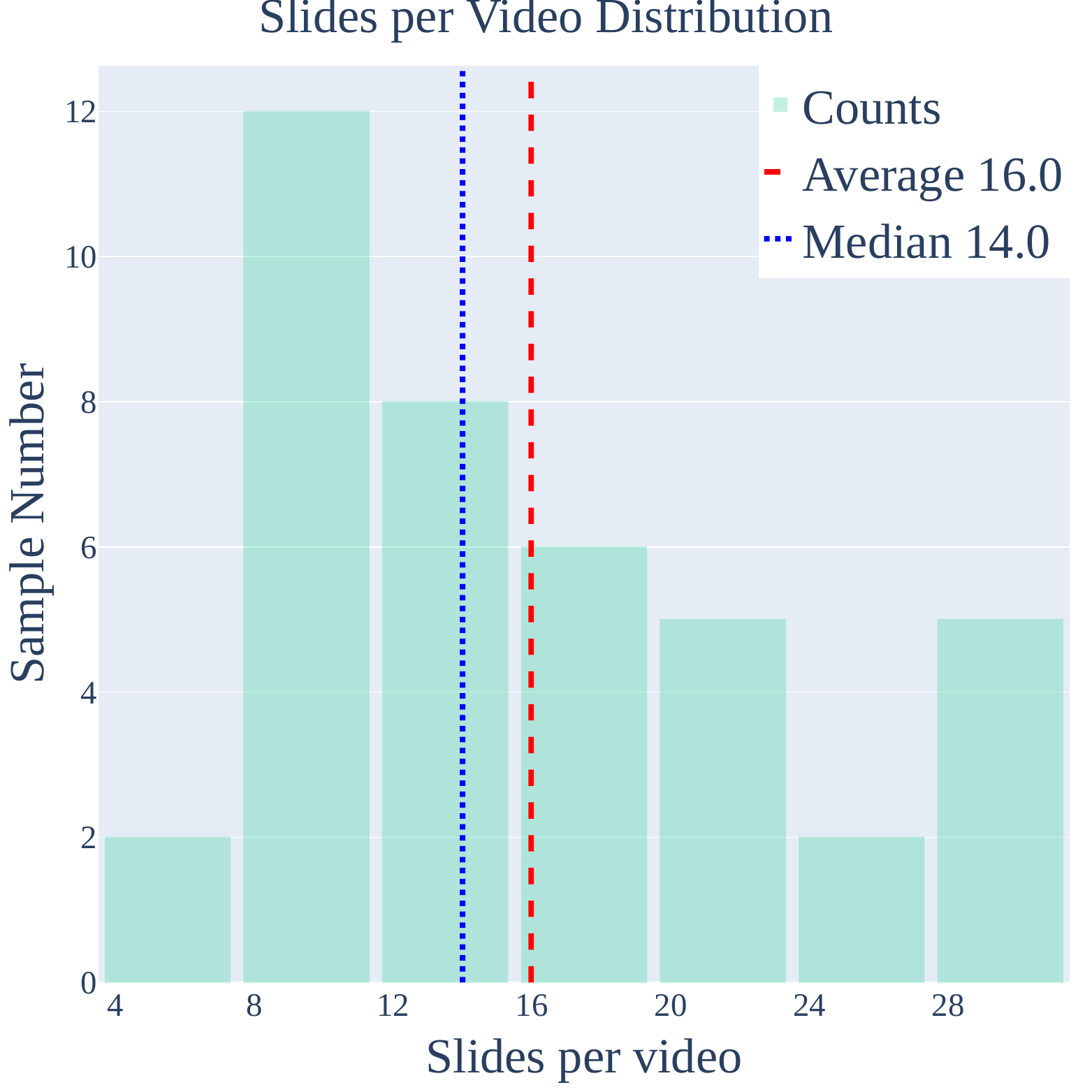}{Slides per video}\hfill
  \sqimg{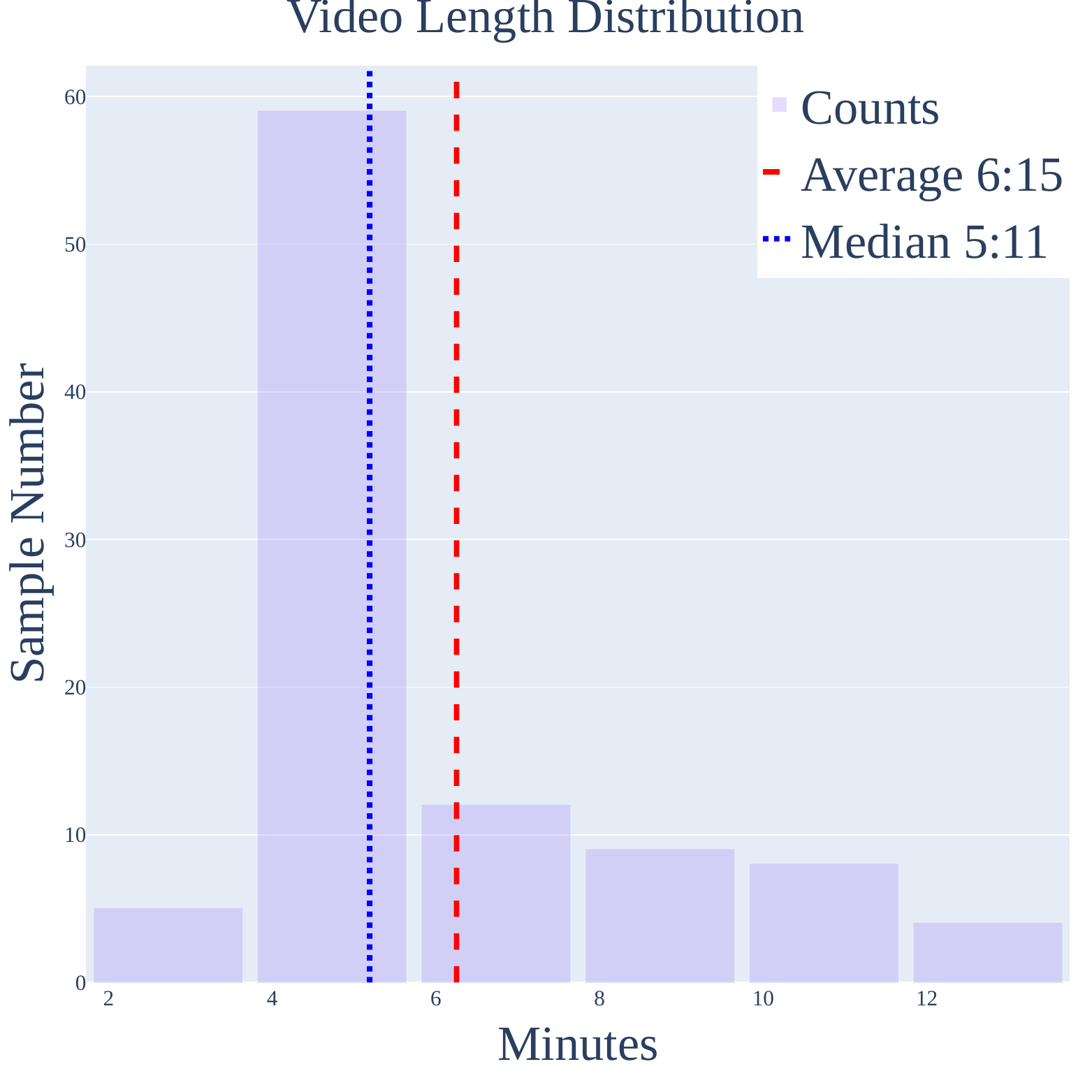}{Video length}

  \caption{\textbf{Statistics of {\bench} benchmark.} It spans diverse topics, with presentations comprising 4--28 slides and lasting 2--14~min, providing a valuable benchmark for the automatic generation and evaluation of academic presentation videos.}
  \label{fig:stat}
\end{figure}
% (\kevin{long-horizon tasks, require multi-turn xxx, thus spend longer time, cost rather than natural vid.})

% video avg len
% slide number 

% Despite 101, we emphasis agentic task rather simple video gen;

% [add figure]
\vspace{-0.5\baselineskip}
\subsection{Evaluation Metrics}
\vspace{-0.8\baselineskip}
Unlike natural video generation, academic presentation videos serve a highly specialized role: they are not merely about visual fidelity but about communicating scholarship. This makes it difficult to directly apply conventional metrics from video synthesis (\textit{e.g.}, FVD, IS, or CLIP-based similarity). Instead, their value lies in how well they \textit{disseminate research, amplify scholarly visibility.}

\begin{figure}[t]
    \centering
    \includegraphics[width=1\linewidth]{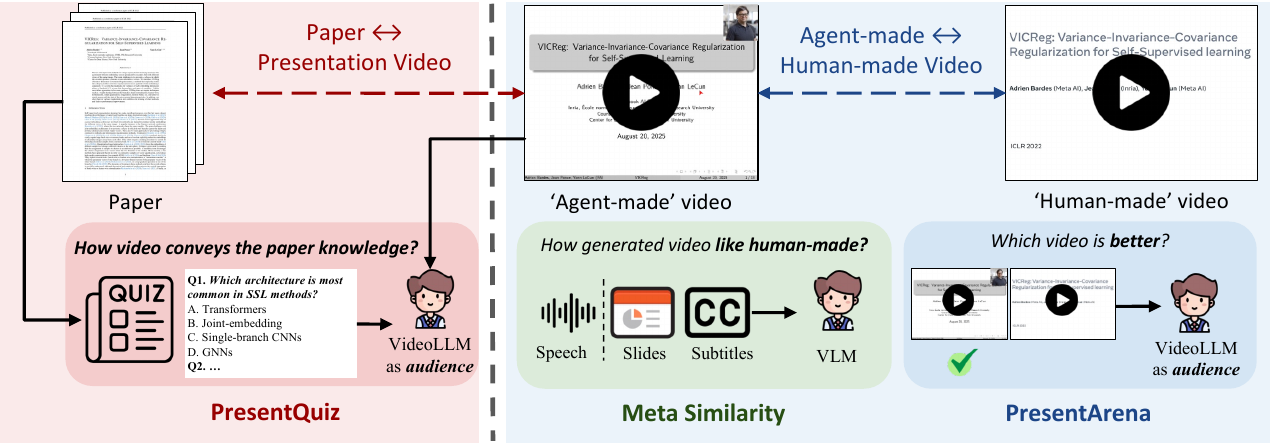}
    \caption{\textbf{Overview of evaluation metrics.} We propose three metrics that systematically evaluate academic presentation video generation from the perspective of the relationship between the generated video and \textbf{(i)} the original paper and \textbf{(ii)} the human-made video.}
    \label{fig:eval}
    \vspace{-0.2\baselineskip}
\end{figure}

% \kevin{compress}
\vspace{-0.2\baselineskip}
From this perspective, we argue that a high-quality academic presentation video should be judged along two complementary dimensions (see Figure~\ref{fig:eval}):
\textbf{For the audience:}  
the video is expected to faithfully convey the paper’s core ideas(\ie~motivation and contributions), while remaining accessible to audiences.
%the video must faithfully convey the paper’s central ideas, such as its motivation, problem formulation, and key contributions; and it should present these ideas in a manner that is easy to follow, allowing viewers from diverse backgrounds to understand the work without being overwhelmed.
\textbf{For the author:} the video should foreground the authors’ intellectual contribution and identity, and enhance the work’s visibility and impact.
To systematically capture these goals, we introduce tailored evaluation metrics specifically designed for academic presentation videos. 
% why we need a pre; what define a good pre;

\vspace{-0.2\baselineskip}
\noindent\textbf{Meta Similarity} \textit{-- How video like human-made?} 
%\kevin{As we have the ground-truth human-made demonstration videos,}
As we have the ground-truth human-made presentation videos with original slides, we evaluate how well the generated intermediate assets (\ie~speech, slides, and subtitles) aligned with the ones created by authors, which serves as the pseudo ground-truth. 
(i) For each slide, we pair the slide image with its corresponding subtitles and submit both the generated pair and the human-made pair to the VLMs to obtain a similarity score on a five-point scale. 
(ii) To further assess speech(\ie~vocal timbre), we uniformly sample a ten-second segment from the presentation audio, encode the generated and human-recorded audio with a speaking embedding model~\cite{audio_embedding}, and compute the cosine similarity between the embeddings to measure speech similarity.

\vspace{-0.2\baselineskip}
\noindent\textbf{PresentArena} \textit{-- Which video is better?} Similar to the human audience watching the presentation, we employ the VideoLLMs as the proxy audience to conduct pairwise comparisons of presentation videos, where the winning rate serves as the metric. For each pair, the model is queried twice in opposite orders: $(A,B)$ and $(B,A)$. This procedure reduces hallucinations and position bias. The two judgments are then aggregated by averaging to obtain a more stable preference estimation.

\vspace{-0.2\baselineskip}
\noindent\textbf{PresentQuiz} \textit{-- How videos conveys the paper knowledge?} Following prior work \cite{pang2025paper2poster}, we evaluate information coverage using a multiple-choice quiz on the presentation video. We first generate a set of questions with four options and the corresponding correct answers from the source paper. Then we ask the VideoLLMs to watch the presentation and answer each question. Overall accuracy serves as the metric, with higher accuracy indicating better information coverage. 

% However, since generated video durations vary and longer videos can trivially include more content, we introduce a length penalty to overlong generations. 
% In practice, consider that researchers usually intend to use concrete video to present the content thus we define a penalized weight $\alpha$ and aim to penalize the generated videos with too long duration: \begingroup
% \small % 可换成 \footnotesize 或 \scriptsize 更小
% % \begin{equation}
% $
% \alpha = \exp\!\left(-\,\frac{\max\{0,\,L^{\mathrm{gen}}-L^{\mathrm{gt}}\}}{L^{\mathrm{gt}}}\right), \tilde{\mathbf{s}} = \mathbf{s}\cdot \alpha,
% $
% % \end{equation}
% % \begin{equation}
% % \tilde{\mathbf{s}} = \mathbf{s}\cdot \alpha.
% % \end{equation}
% \endgroup
% where the  $L^{\mathrm{gt}}$ and $L^{\mathrm{gen}}$ denote the ground-truth and generated video durations, and let $s$ be the original accuracy score. 
% The penalized score, $\tilde{s}$, implies that a good presentation video should try to convey paper's information within limited duration.

\vspace{-0.3\baselineskip}
\noindent\textbf{IP Memory} \textit{-- How videos affect the author's visibility and work impact?} Another key purpose of academic presentation videos is to enhance the visibility and impact of the author’s work. Yet, this metric is unclear and difficult to simulate and thus remains an open problem. In real-conference settings, audiences who recall a scholar after attending their presentation are more inclined to pose relevant questions in later interactions. Motivated by this phenomenon, we propose a metric to assess how effectively a presentation video enables the audience to recall the work. Additional implementation details are provided in Appendix~\ref{sec:ip}.

\vspace{-0.3\baselineskip}
\noindent Furthermore, to ablate the contribution of each component, we evaluate both the quality and the gains provided by individual components (\eg~slides, cursor, and presenter). Notably, to further assess presentation videos from the user perspective, we conduct human studies to evaluate the results.

\vspace{-0.8\baselineskip} 
\section{\agent~Agent}
\begin{figure}[t]
    \centering
    \includegraphics[width=\linewidth]{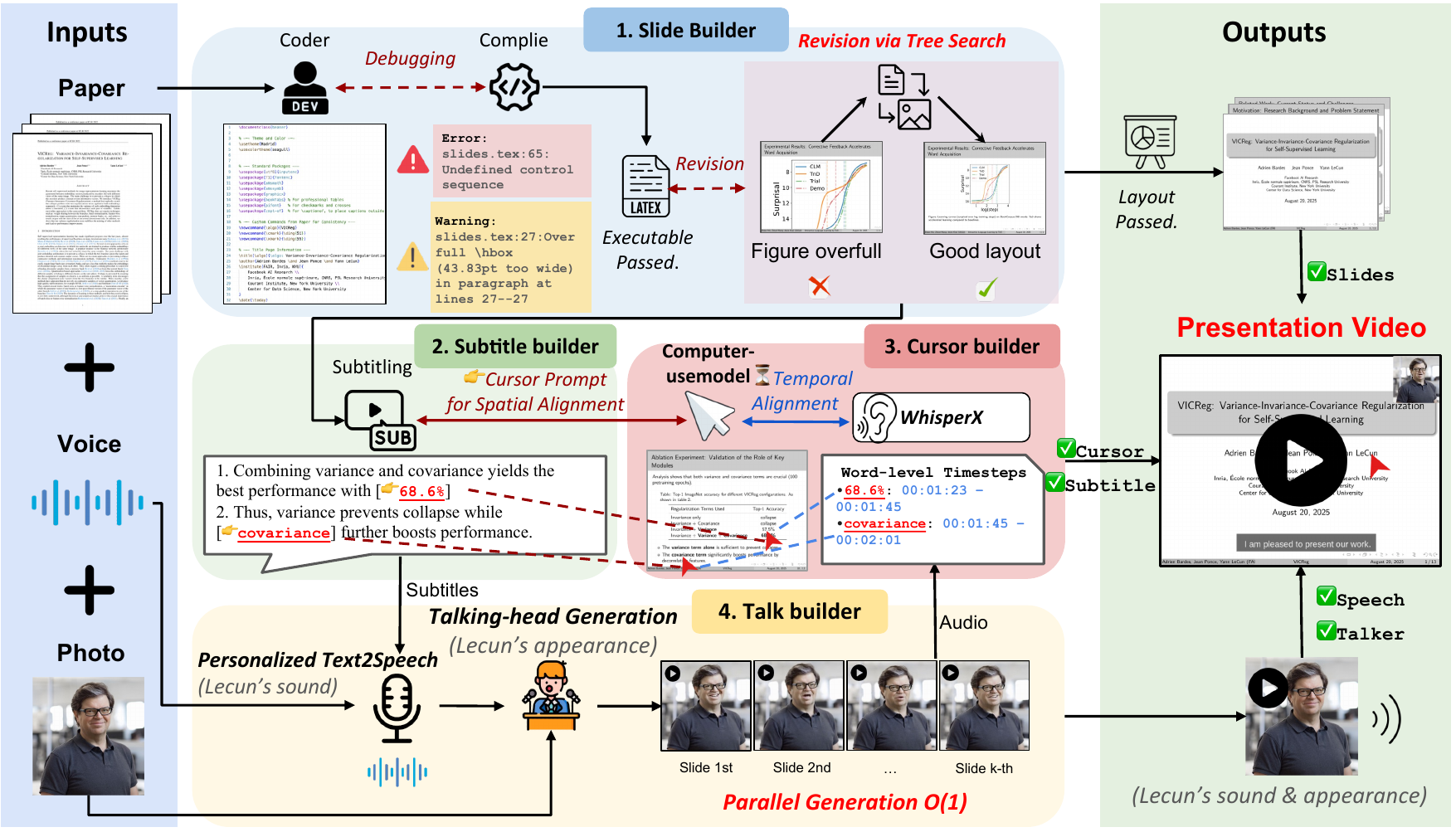}
    \caption{\textbf{Overview of {\agent}.} Our pipeline comprises three key modules: \textbf{(i)} tree search visual choice for fine-grained slide layout optimization; \textbf{(ii)} a GUI-grounded model paired with WhisperX for spatiotemporally aligned cursor grounding; and \textbf{(iii)} slide-wise parallel generation for efficiency.}
    \label{fig:method}
\end{figure}

\vspace{-0.8\baselineskip} 
\textbf{Overview.} 
To address these challenges and liberate researchers from the burdensome task of manual video preparation, we introduce {\agent}, a multi-agent framework designed to automatically generate presentation videos directly from academic papers. 
As illustrated in Figure~\ref{fig:method}, to decouple the different roles, making the method scalable and flexible, the pipeline comprises four builders:
% \kevin{pdf /pptx I/O should be acceptable, here don't include quite a lot of details such as GUI / WhisperX}
\textbf{(i)} Slide builder. Given the paper, we first synthesize slides with \LaTeX~code and refine them with compilation feedback to correct grammar and optimize layout; 
%’s \LaTeX{} project (\textit{e.g.}, source code and figures)
\textbf{(ii)} Subtitle builder. The slides are then processed by a VLM to generate subtitles and sentence-level visual-focus prompts;  \textbf{(iii)} Cursor builder. These prompts are then grounded into on-screen cursor coordinates and synchronized with the narration.
% The finalized slides are then processed by a vision–language model (VLM) to produce a subtitle script and assign a visual-focus cue to each sentence. Then, a GUI model grounds every cue as a precise on-screen cursor trajectory, while WhisperX temporally aligns the cursor with the narration. 
\textbf{(iv)} Talker builder. Given the voice sample and the portrait of the speaker, text-to-speech and talking-head modules generate a realistic, personalized talker video. 
%For clarity, we denote $p,a,v$ for the slides, speech audio, and human presentation video, respectively, and $l,a_o,v_o$ for the input \LaTeX{} project, the speaker’s voice sample, and portrait.
For clarity, we denote the paper document, author portrait, and voice sample as $\mathcal{D}$, $\mathcal{I}$, and $\mathcal{A}$, respectively.

\vspace{-0.4\baselineskip} 

\vspace{-0.5\baselineskip}
\subsection{Slide Builder}
\vspace{-0.5\baselineskip}
% \kevin{accept $\mathcal{D}$, output $\mathcal{S}$, $\mathcal{S}_0$ per page}
A prerequisite for producing a presentation video is the creation of the slides. Despite there being some existing works~\cite{zheng2025pptagent}, we target the generation of academic slides with fine-grained layouts and formal structure from scratch.
% \textbf{Why we use Latex \& Beamer}
Rather than selecting a template and iteratively editing it with VLMs, we generate slides directly from a paper’s \LaTeX{} project by prompting the model to write Beamer code. We adopt Beamer for three reasons: \textbf{(i)} \LaTeX{}’s declarative typesetting automatically arranges text block and figures from their parameters without explicitly planing the positions; \textbf{(ii)} Beamer is compact and expressive, representing the same content in fewer lines than XML-based formats; and \textbf{(iii)} Beamer provides well-designed, formally configured styles (\textit{e.g.}, page numbers, section headers, hyperlinks) that are well suited to academic slide design.

% As \(l=(t,\mathcal{F})\) denote the project of paper, where \(t\) is the text source code and \(\mathcal{F}=\{f_i\}_{i=1}^{n}\) are the figure paths.
Given the paper $\mathcal{D}$ as input, the LLM first produces a draft slide code. We compile this code to collect diagnostics(\ie~errors and warnings). Then, we use the error information to elicit a repaired correct code.
% , as shown in Alogrithm~\ref{alg:compile-repair}
This procedure ensures that the generated Beamer code is grammatically correct and effectively leverages and faithfully covers its content.
% , yielding the content of slides that are ready to use.

Although \LaTeX{} can automatically arrange the location of the contents in the slides, the generated slides could sometimes still suffer from inappropriate layouts (\textit{e.g.}, overflow) due to the unsuitable parameters for figure or text font size. However, as the compilation warning signals potential layout issues, we are able to first use them to identify the slides that require refinement.
% \kevin{One challenging is that LLm fail to preceive visually, thus} 
% as LLM fail to perceive visual feedback like human designers,
\begin{wrapfigure}[18]{r}{0.4\textwidth} % 12=大致占用的行数，可调 8~16
  %\vspace{-0.5\baselineskip}              % 往上顶，避免与上一行留白
  \centering
  \includegraphics[width=\linewidth]{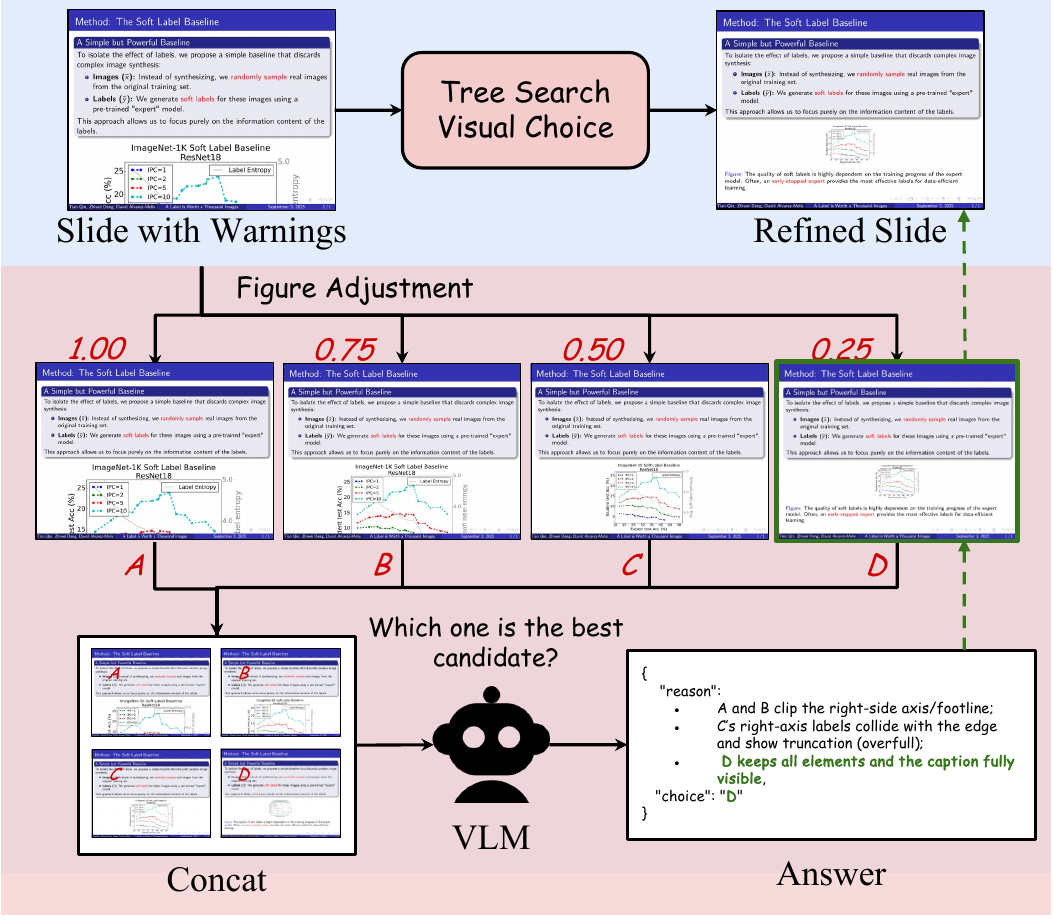}
  \captionsetup{skip=2pt}                 % 图↔标题更紧
  \caption{\textbf{Tree Search Visual Choice.} It combines a rule-based proposal mechanism with VLM-based scoring to select the optimal candidate.}
  \label{fig:mcts}
  %\vspace{-1\baselineskip}              % 与后续正文再贴近
\end{wrapfigure}

\vspace{-1\baselineskip}
\noindent\textbf{Tree Search Visual Choice.} After localizing the slides that require refinement, the key challenge is how to adjust their layouts effectively. As LLMs/VLMs fail to perceive real-time visual feedback like human designers, we observe that prompting the them to directly tune numeric layout parameters (\textit{e.g.}, font sizes, margins, figure scales) is ineffective: the models are largely \textbf{insensitive} to small numeric changes, yielding unstable and inefficient refinement, consistent with limitations of the parameter-editing strategy in PPTAgent~\cite{zheng2025pptagent}. To address this limitation, we introduce a \emph{visual-selection} module for overflowed slides. The module first constructs the neighborhoods of layout variants for the current slide by rule-based adjusting the figure and text parameters, renders each variant to an image, and then uses the VLMs as a judge to score the candidates and select the one with the best layout. Specifically, for text-only slides, we sweep the font size; for slides with figures, we first vary the figure scaling factors (\textit{e.g.}, 1.25, 0.75, 0.5, 0.25) and then reduce the font size, details shown in Figure~\ref{fig:mcts}. These edits are straightforward in \LaTeX{} Beamer, whose structured syntax automatically reflows content as parameter changes. This module \textbf{decouples discrete layout search from semantic reasoning} and reliably resolves overflow cases with minimal time and tokens.

After fixing the errors and adjusting the parameters, we compile the slide code to obtain the finalized slides $\mathcal{S}_i,i=1,\dots,n$ with fine-grained layouts, where $n$ indicates the number of slides.
% \kevin{wondering add a small fig here. add a figure}

% \textbf{decompose, modular}
% either pdf input / latex input are both available;

% non-trivial
% \begin{wrapfigure}[12]{r}{0.5\textwidth} % 12=大致占用的行数，可调 8~16
%   %\vspace{-0.5\baselineskip}              % 往上顶，避免与上一行留白
%   \centering
%   \includegraphics[width=\linewidth]{figure/tree_search.pdf}
%   \captionsetup{skip=2pt}                 % 图↔标题更紧
%   \caption{\textbf{Tree Search Visual Choice.} It combines a rule-based proposal mechanism with VLM-based scoring to select the optimal candidate.}
%   \label{fig:mcts}
%   %\vspace{-1\baselineskip}              % 与后续正文再贴近
% \end{wrapfigure}

\vspace{-0.5\baselineskip} 
\subsection{Subtitle Builder}
\vspace{-0.5\baselineskip} 
% \kevin{$\mathcal{S}_0$ slide, output a transcript $\mathcal{T}_0$, cursor prompt $\mathcal{P}_{0}^i$}
% \kevin{what's the challenging here, are we apply the parallel gen here also? why slide in rather than paper in? because less context; why create visual-focus from subtitle? usually, when human talking some slide with sentence, they have some focusness on keywords; \textbf{inputs, outputs}}
As the speech should follow the slides, given the generated slide $\mathcal{S}_i$, we rasterize them into images and pass them to a VLM, which produces sentence-level subtitles $T_{i}^{j}$ and its corresponding visual-focus prompt $P_{i}^{j}$. The visual-focus prompt serves as an intermediate representation linking speech to the cursor, enabling precise temporal and spatial alignment of the cursor with the narration in order to improve audience guidance, which will be discussed in Section~\ref{sec:cursor}.

\vspace{-0.5\baselineskip} 
\subsection{Talker Builder}
\vspace{-0.5\baselineskip} 
% consider formal, gesture, thus we choose which models (digital)
% \textbf{Key challenges -- bottlenect of cascade generation long video}
% 5min, per slide very slow;
% [highlight insights from real huamn recording]
% usually / practically, gen per page independent, so this motivate us to develop a parallel strategy,
% \kevin{$\mathcal{T}$, $\mathcal{A}$, $\widetilde{\mathcal{A}}$, $\mathcal{V}_0$}
The presenter video is vital for audience engagement and conveying the researcher’s scholarly identity (\textit{e.g.}, face and voice). Given the subtitles $\mathcal{T}_i$, the author’s portrait $\mathcal{I}$, and a short voice sample $\mathcal{A}$, our objective is to synthesize a presenter video that delivers the slide content in the author’s voice, with faithful identity preservation and lip–audio synchronization.

\vspace{-0.4\baselineskip}
\noindent\textbf{Subtitle-to-Speech.} Given subtitles and voice sample, we use F5-TTS~\cite{tts-f5} to generate speech audio per slide, \begingroup \small $
\widetilde{\mathcal{A}}_i = \operatorname{TTS}\!\left( \{\, T_{i}^{j} \,\}_{j=1}^{m_i} |\mathcal{A}\right), i=1,\ldots,n, $
\endgroup ~where $m_i$ is the number of the sentences in $\mathcal{T}_i$.

\vspace{-0.4\baselineskip}
\noindent\textbf{Parallel Talkinghead Generation.} 
% \kevin{add intuition here, in real-world human also prepare videos like this} 
To balance fidelity and efficiency, we use Hallo2~\cite{cui2024hallo2} for head-only synthesis and employ FantasyTalking~\cite{fantasytalking} to support talking generation with upper-body articulation.
%considering its higher computational cost. 
A persistent challenge is the long generation time: generating only a few minutes of talking-head video typically takes several hours, and some models(\textit{e.g.}, FantasyTalking) do not yet natively support long-video generation. 
% To alleviate this bottleneck, 
Inspired by the common practice of slide-by-slide recording and the independence between each slide, we synthesize the presenter video on a per-slide basis. Specifically, for each slide $\mathcal{S}_i$, given the audio condition $\widetilde{\mathcal{A}}_i$ and portrait $\mathcal{I}$, we generate an independent clip $\mathcal{V}_i$ and execute these jobs in parallel, markedly reducing generation time: \begingroup \small $\mathcal{V}_i = \mathcal{G}\!\left(\widetilde{\mathcal{A}}_i, \mathcal{I}\right), i=1,\ldots,n,$\endgroup ~where $\mathcal{G}$ represents the talking-head generation model.
This design is justified because slide transitions are hard scene changes, and the temporal continuity of the presenter across adjacent slides is unnecessary.

\vspace{-0.5\baselineskip} 
\subsection{Cursor Builder}
\vspace{-0.5\baselineskip} 
\label{sec:cursor}
% \textbf{Key challenges -- How to associate / map speech improtance with visually}
% usually, the speaker will have keywords in each sentence, so parse keyword, then, motivated by strong Computer-use model, this can be a query XX
% [fig.] how speech sentence -> keyword -> computer use [x,y] -> screenshot
\textbf{Spatial-Temporal Grounding.}
% \kevin{input $\mathcal{P}_{0}^i$ with a slide $\mathcal{S}_0$, then output a coordinate $[x,y]$}
% \kevin{For spatial: we are motivated by Computer-ues grounding, which simulates user interaction xxx}
In practice, presenters leverage the cursor as an attentional guide: a well-aligned cursor trajectory minimizes extraneous cognitive load, helps the audience track the presentation, and keeps focus on the key content. However, automatic cursor-trajectory grounding is nontrivial, requiring simultaneous alignment to the timing of speech and the visual semantics of the slides. To simplify the task, we assume that the cursor will stay still within a sentence and only move between the sentences. Thus, we estimate a per-sentence cursor location and time span. For spatial alignment, motivated by strong computer-use models~\cite{lin2025showui,qin2025ui} which simulate user interaction with the screenshot, we propose to ground the cursor location $(x,y)$ for each sentence with the visual focus prompt $\mathcal{P}_{i}^{j}$ by UI-TARS~\cite{qin2025ui}. To achieve precise temporal alignment, we then use WhisperX~\cite{bain2023whisperx} to extract word-level timestamps and align them with the corresponding sentence in the subtitles to derive the start and end times $(t_s,t_e)$ of each cursor segment.
% \kevin{For temporal: we aim to get precise alignment, while sentence is not satisfy, thus we use world-alignment}

% 1. human-made video
% 2. paper
% Wan2.1
% PPTAgent

% pptx / beamer / beamer + debug / bearm +debug+tree search

% user study

% IP: ask a reasonable question

\vspace{-0.5\baselineskip} 
\section{Experiments}
\vspace{-0.5\baselineskip} 
\subsection{Baseline and Settings}
\vspace{-0.5\baselineskip} 
We evaluate three categories of baselines: \textbf{(i)} End-to-end Methods~\cite{wan,deepmind2025veo3}, where natural video generation models produce the presentation video directly from a prompt generated by paper; \textbf{(ii)} Multi-Agent Frameworks~\cite{shi2025presentagent,zheng2025pptagent}, which combine slide generation with text-to-speech generation and compose them into a presentation video; and \textbf{(iii)} {\agent}, our method and its variants. For the VLM and VideoLLM, we choose \textit{GPT-4.1} and \textit{Gemini-2.5-Flash}, respectively, for a favorable efficiency and performance trade-off. We perform inference using eight NVIDIA RTX A6000 GPUs.

% End-to-end, multi-agent, slide-agent, ours
% \vspace{-1\baselineskip} 
\vspace{-0.5\baselineskip} 
\subsection{Main Results}
% \kevin{highlight exps as showui}
\vspace{-0.5\baselineskip} 
\textbf{Meta Similarity.} We evaluate the alignment of the generated slides, subtitles, and speech with corresponding human-authored ones. For speech, we randomly sample a 10-second audio segment from the video generated by each method and compute the cosine similarity between its embeddings~\cite{audio_embedding} and those of the author’s speech. As shown in Table~\ref{tab:main_result}, {\agent} attains the highest scores in both speech and content similarity, demonstrating that its outputs \textbf{align most closely with human creation} among all baselines. We attribute this performance to personalized TTS and our slide-generation design: \textbf{(i)} adopting \textit{Beamer}, which provides formal, academically styled templates while \LaTeX{} automatically arranges content within each slide; and \textbf{(ii)} a tree search visual choice layout refinement that further enforces fine-grained slide layouts as commonly observed in human-authored slides.

\vspace{-0.4\baselineskip} 
\textbf{PresentArena.} We compare the presentation videos generated by each method against the human-made videos. As an automatic evaluator, we prompt the VideoLLMs as a judge to determine which presentation is better with respect to clarity, delivery, and engagement. As shown in Table~\ref{tab:main_result}, {\agent} attains the highest pairwise winning rate among all baselines, indicating that our method produces presentation videos with \textbf{superior overall perceived quality}. Notably, {\agent} outperforms its variants without the talker and cursor by 1.8\%, highlighting the gains introduced by these components and implying that the VideoLLM \textbf{favors presentation videos with a talker presenting}.

% \kevin{\textbf{Talkinghead} XXXX}

\vspace{-0.4\baselineskip} 
\textbf{PresentQuiz.} To assess information coverage, we conduct a VideoQA evaluation. Following prior work on posters~\cite{pang2025paper2poster}, we construct QA sets by prompting an LLM to generate questions targeting (i) fine-grained details and (ii) higher-level understanding of the paper. The videos and QA sets are then fed into a VideoLLM to conduct the quiz.
As shown in Table~\ref{tab:main_result}, {\agent} achieves superior performance across both aspects, outperforming HumanMade and PresentAgent despite shorter video length. This indicates that {\agent} produces videos that are \textbf{more informative within shorter durations}. Furthermore, the absence of the talker or cursor results in performance degradation, as the cursor trajectory  potentially \textbf{guides the attention and supports accurate grounding of the key contents} for the VideoLLMs during inference, referring to Table~\ref{table:cursor} for more details.

\vspace{-0.4\baselineskip} 
\textbf{IP Memory.} We evaluate the degree to which the generated presentation videos facilitate audience retention of the work, thereby assessing their memorability and lasting impact. {\agent} achieves the highest recall accuracy. This improvement mainly stems from the inclusion of an engaging talker with the author’s figure and voice, which \textbf{significantly helps the audience retain the video content}.

% {\agent}’s gains stem from our slide-generation design and our principled cursor-trajectory synthesis, which together guide attention and enable accurate grounding of key content for the VideoLLMs. Although PresentAgent attains slightly higher unpenalized accuracy, its length-penalized score is markedly lower owing to the excessive duration of its generated videos compared with human-made videos.

\begin{wrapfigure}[14]{r}{0.32\textwidth} % 12=大致占用的行数，可调 8~16
  \vspace{-0.8\baselineskip}              % 往上顶，避免与上一行留白
  \centering
  \small
  \includegraphics[width=\linewidth]{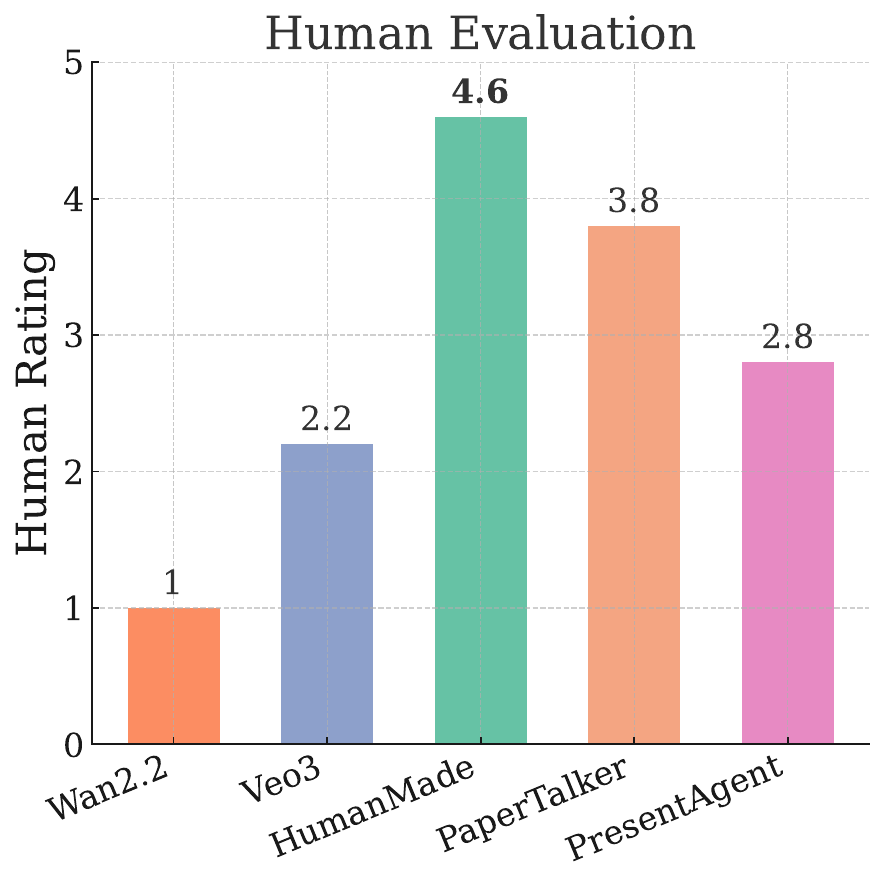}
  \captionsetup{skip=0pt}                 % 图↔标题更紧
  \caption{\footnotesize \textbf{Human evaluation}. We randomly sample the generated results from ten papers for evaluation.}
  \label{fig:human_eval}
  %\vspace{-1\baselineskip}              % 与后续正文再贴近
  %\vspace{-0.5\baselineskip}
\end{wrapfigure}
\vspace{-0.3\baselineskip} 
\textbf{Human Evaluation.}
To further assess the quality of the generated presentations from the user perspective, we conducted a human evaluation in which ten participants were provided with each paper along with its corresponding presentation videos generated by different methods. Participants were asked to rank the videos according to their preferences($1\text{(worse)}-5\text{(best)}$).
As shown in Figure~\ref{fig:human_eval}, human-made videos achieve the highest score, with {\agent} ranking second and outperforming all other baselines. This demonstrates that presentation videos generated by {\agent} \textbf{gain consistently favor from human users} over other baselines and \textbf{comparable to human-made}.

\vspace{-0.3\baselineskip} 
\textbf{Efficiency Analysis.} As shown in Table~\ref{table:cost}, {\agent} achieves the lowest cost. This efficiency stems from our slide-generation design: adopting \textit{Beamer} \textbf{reduces token usage} for slide creation, and our tree search visual choice module which is a \textbf{lightweight post-processing step}. Runtime is further reduced by our parallel talking-head generation mechanism. By contrast, PresentAgent~\cite{shi2025presentagent} incurs higher token costs due to frequent refinement queries during slide editing.

\begin{table*}[t]
  \centering
  \setlength{\tabcolsep}{4pt}
  \footnotesize
  \caption{\textbf{Detailed evaluation result of {\bench} across three baselines.} $\text{PaperTalk}^{*}$ represents a simple version without presenter and cursor. \textbf{Bold} and \underline{Underline} indicates the best and the second. NA means the results of method are not applicable to the metric.}
  \label{tab:main_result}

  \begin{tabular*}{\linewidth}{l c c c c c c c}

    \toprule
    \multirow{2}{*}{Method} &
    \multicolumn{2}{c}{Similarity$\uparrow$} &
    \multirow{2}{*}{Arena$\uparrow$} &
    \multicolumn{2}{c}{PresentQuiz Acc.$\uparrow$} &
    \multirow{2}{*}{IP Memory$\uparrow$} &
    \multirow{2}{*}{Avg. Duration(s)}\\
    \cmidrule(lr){2-3}\cmidrule(lr){5-6}
    & Speech & Content & & Detail & Under.\\
    \midrule
    HumanMade & \textbf{1.00} & \textbf{5.00}  & \textbf{50.0\%}  & 0.738  & 0.908  & - & 375.15\\
    \midrule
    \rowcolor{gray!20} Wan2.2~\cite{wan} & NA & NA & 1.1\% & 0.251 & 0.551 & 11.5\% & 4.00\\
    \rowcolor{gray!20} Veo3~\cite{deepmind2025veo3} & 0.133 & NA & 1.2\% & 0.367 & 0.585 & 31.3\% & 8.00\\
    PresentAgent\textsubscript{QWEN}~\cite{shi2025presentagent} & - & 0.24 & - & - & - & - & - \\
    PresentAgent\textsubscript{GPT4.1}~\cite{shi2025presentagent} & 0.045 & 1.47 & 2.0\% & 0.548 & 0.654 & 12.5\% & 430.20\\
    \midrule
    PaperTalk\textsubscript{QWEN} & - & 1.66 & - & - & - & - & - \\
    $\text{PaperTalk\textsubscript{GPT4.1}}^{*}$ & \underline{0.646} & \underline{1.97} & 15.2\% & \underline{0.835}  & \underline{0.949} & \underline{37.5\%} & 234.36 \\
    PaperTalk\textsubscript{GPT4.1} & \underline{0.646} & \underline{1.97} & \underline{17.0}\% & \textbf{0.842}  & \textbf{0.951} & \textbf{50.0\%} & 234.36\\
    \bottomrule
  \end{tabular*}
\end{table*}
\vspace{-0.4\baselineskip}

\vspace{-0.5\baselineskip} 
\subsection{Qualitative Analysis}
\begin{figure}
    \centering
    \includegraphics[width=\linewidth]{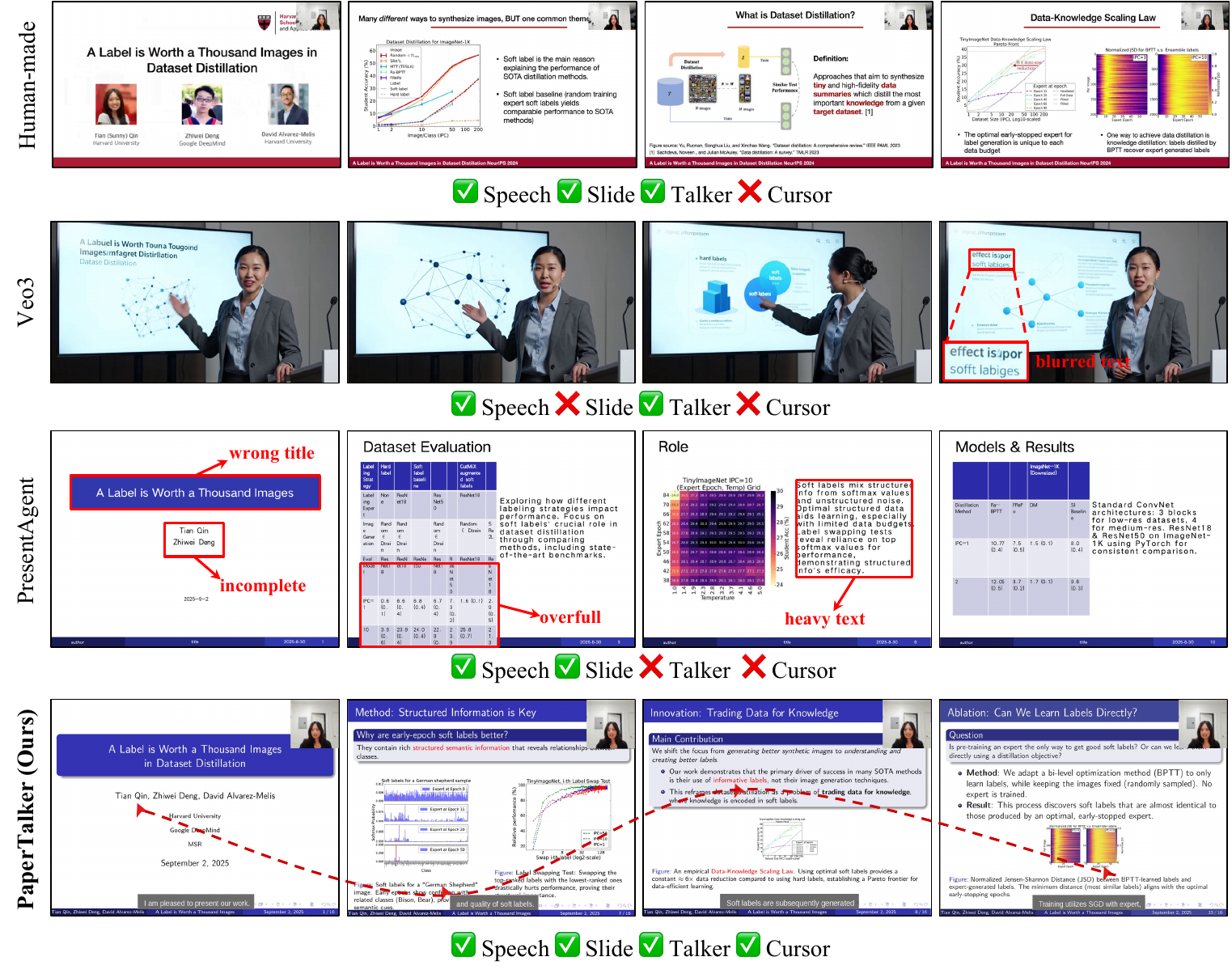}
    \captionsetup{skip=0pt}  
    \caption{\textbf{Visualization of generated results.} {\agent} produces presentation videos with rich, fine-grained slide content, accurate cursor grounding, and an engaging talker; in contrast, Veo3~\cite{deepmind2025veo3} yields blurred text and incomplete information coverage, while PresentAgent~\cite{shi2025presentagent} produces text-heavy slides and suffers from overfull layout issues and inaccurate information (\textit{e.g.}, title and institutions).}
    \label{fig:compare}
    \vspace{-0.2\baselineskip} 
\end{figure}

\vspace{-0.5\baselineskip} 
\begin{wraptable}[7]{r}{0.55\textwidth}
  \vspace{-1.5\baselineskip} 
  \centering
  \tablesize
  \caption{\textbf{Generation cost for each method.}}
  {
  \setlength{\tabcolsep}{3pt}
  \begin{tabular}{lccc}
    \toprule
    Method & Token (K) $\downarrow$ & Time (min.)$\downarrow$ & Cost (\text{\$})$\downarrow$ \\
    \midrule
    Veo3~\cite{deepmind2025veo3} & NA & \textbf{0.4} & 1.667 \\
    Wan2.2~\cite{wan} & NA & 8.1 & 0.280 \\
    \hline
    PresentAgent~\cite{shi2025presentagent} & 241 & 39.5 & 0.003  \\
    PaperTalker (w/o Talker) &  \textbf{62} & 15.6 & \textbf{0.001}\\    
    PaperTalker (w/o Par.) & \textbf{62} & 287.2 & \textbf{0.001}\\
    PaperTalker & \textbf{62} & 48.1 \textbf{(6$\times$)} & \textbf{0.001} \\
    \bottomrule
  \end{tabular}
  }
  \label{table:cost}
  \vspace{-2\baselineskip} 
\end{wraptable}
As shown in Figure~\ref{fig:compare}, {\agent} produces presentation videos that \textbf{most closely align with the human-made} ones. While Veo3~\cite{deepmind2025veo3} renders a high-quality speaker in front of the screen, it is constrained by short duration (\textit{e.g.}, 8s) and blurred text. Besides, PresentAgent\cite{shi2025presentagent} typically suffers from the absence of the presenter and slide-design errors (\textit{e.g.}, overflow, incorrect title, incomplete author lists, and institutions).

% table1 comparion, similarity(audio, content), video qa acc, acadimact IP

% \usepackage{booktabs}
% \usepackage{multirow}

% ablation: cursor accuary

% 不同的LLM/VLM
% Layout refinement 
%1) multi-turn 
%2) single turn (single/multi)
% \vspace{-1\baselineskip} 
\vspace{-0.5\baselineskip} 
\subsection{Key Ablations}
\vspace{-0.5\baselineskip} 
%\noindent\textbf{Is Talking Head Necessary?} In presentation videos, the presence of a presenter substantially affects audience engagement. Empirically, the PresentArena results in Table~\ref{tab:main_result} show {\agent} outperforming its no-presenter, no-cursor variant PaperTalker* by more than 10\%, implying that the VideoLLM prefers presentation videos which includes a presenter. 
%\kevin{PaperTalker*} \kevin{add exactly number}
% \kevin{Human eval;}
\begin{wraptable}[3]{r}{0.4\textwidth} % 右侧；宽度可调；[6] 是预估占用行数
  \vspace{-2\baselineskip}               % 可选：向上收一点
  \tablesize
  \centering
    \caption{\textbf{Ablation study on cursor.}}
  \begin{tabular}{lc}
    \toprule
    Method & Accuracy$\uparrow$ \\
    \midrule
    PaperTalker (w/o Cursor) &  0.084 \\
    PaperTalker &\textbf{0.633} \\
    \bottomrule
    \label{table:cursor}
  \end{tabular}
\end{wraptable}
\noindent\textbf{What benefits are brought by Cursor Highlight?} 
Motivated by the observation that a cursor typically helps audiences locate the relevant region, we hypothesize that a visible cursor, by providing an explicit spatial cue, facilitates content grounding for VLMs. To evaluate this, we design a localization QA task: for each subtitle sentence and its corresponding slide, a VLM generates a four-option multiple-choice question about the sentence’s corresponding position on the slide. The VLMs are then prompted to answer using slide screenshots, with or without the cursor, and accuracy is measured as the metric. As shown in Table~\ref{table:cursor}, the accuracy is much higher with the cursor highlight, corroborating its \textbf{importance for the audience's visual grounding accessibility} of presentation videos. 

\begin{wraptable}[8]{r}{0.6\textwidth} % [10] 行数可调, r=右侧, 宽度自己调
 \vspace{-1\baselineskip} 
  \centering
  \tablesize
  \caption{\textbf{Evaluation result on slide quality}.}
  {
  \setlength{\tabcolsep}{1pt} % 默认是6pt，这里改小
  \begin{tabular}{lccc}
    \toprule
    Method & Content ($\uparrow$) & Design ($\uparrow$) & Coherence ($\uparrow$) \\
    \midrule
    HumanMade & \textbf{4.43} & \textbf{2.85} & 2.73 \\
    \midrule
    $\text{PPTAgent}_\text{Qwen7B}$~\cite{zheng2025pptagent} & 3.43 & 1.57 & 1.29 \\
    $\text{PaperTalker}_\text{Qwen7B}$ & 4.00 & 2.53 & \underline{3.11} \\
    \midrule
    $\text{PPTAgent}_\text{GPT4.1}$~\cite{zheng2025pptagent} & 4.07 & 2.02 & 2.06 \\
    $\text{PaperTalker}_\text{GPT4.1}$(w/o Tree Search) & 4.33 & \underline{2.73} & \textbf{3.84} \\
    $\text{PaperTalker}_\text{GPT4.1}$ & \underline{4.34} & \textbf{2.85} & \textbf{3.84} \\
    \bottomrule
  \end{tabular}
  }
  \label{tab:slide_quality}
\end{wraptable}
\noindent\textbf{How does tree search visual choice improve slide quality?} To assess the contribution of the tree-search visual choice module, we conduct an ablation experiment, as shown in Table~\ref{tab:slide_quality}. In line with prior work on slide generation~\cite{zheng2025pptagent}, we assess the generated slides using a VLM on a 1–5 scale across content, design, and coherence. The results show a pronounced decline in design quality when layout refinement is removed, highlighting \textbf{the tree-search visual choice module as a key component for slide creation} (\ie~resolving overfull issues), referring to Figures~\ref{fig:tree_vis} for visualization.

%\kevin{same as aboveww}
\vspace{-0.5\baselineskip} 
\section{Conclusions}
\vspace{-0.5\baselineskip} 
This work tackles the long-standing bottleneck of presentation video generation by agent automation. 
With \bench, we provide the first comprehensive benchmark and well-designed metrics to rigorously evaluate presentation videos in terms of quality, knowledge coverage, and academic memorability. 
Our proposed \agent~framework demonstrates that automated generation of ready-to-use academic presentation videos is both feasible and effective, producing outputs that closely approximate author-recorded presentations while significantly reducing production time by 6 times. 
We hope our work advances AI for Research and supports scalable scholarly communication.

\newpage
\bibliographystyle{plainnat}
\bibliography{main} 

\newpage
\newcommand{\AppendixSetup}{%
  \appendix
  \renewcommand{\thesection}{\Alph{section}} % A/B/C...
  \setcounter{tocdepth}{2}\setcounter{secnumdepth}{2}

  % 章节标题样式
  \titleformat{\section}{\large\bfseries\color{red}}{\thesection}{1em}{}
  \titleformat{\subsection}{\normalsize\bfseries}{\thesubsection}{0.8em}{}

  % 附录目录条目样式（点线 + 右对齐页码）
  \titlecontents{section}
    [0em]{\bfseries\color{red}}{\contentslabel{1.2em}}{}{\titlerule*[.5pc]{.}\contentspage}[\vspace{0.2em}]
  \titlecontents{subsection}
    [1.8em]{}{\contentslabel{1.6em}}{}{\titlerule*[.5pc]{.}\contentspage}[\vspace{0.1em}]
}

\newcommand{\AppendixTitlePage}{%
  \begin{center}
    \LARGE\bfseries Appendix\par
  \end{center}
  \vspace{0.75em}
  \noindent\textbf{Contents}\par
  \vspace{0.5em}
}

% ===== 第一页只放目录 =====
\clearpage
\AppendixSetup
\startcontents[app]          % 开始捕获“附录目录”
\AppendixTitlePage
\printcontents[app]{l}{1}{}  % 打印仅附录目录（section+subsection）
\clearpage                   % ← 目录页结束；正文从新页开始

\section{Evaluation Metrics}
\subsection{IP Memory}
\label{sec:ip}
We propose a novel metric to evaluate how well an audience retains a work after watching its presentation video. Motivated by \textbf{real-world conference interactions}, this metric assesses whether an audience member, after viewing several presentation videos, can recall the work and pose a relevant question when meeting the author.

To operationalize this, we construct video–question pairs by sampling a five-second clip from each presentation video and selecting a corresponding understanding-level question from PresentQuiz. A VideoLLM serves as the audience proxy: it is presented with four randomly sampled video–question pairs, where the videos and questions are shuffled, together with an image of one speaker as the query. The model is then asked to identify the relevant question to pose, and the accuracy quantifies the IP Memory score. Higher recall accuracy indicates that the generated results are more impressive and hold greater potential for lasting impact.

\section{Experiment}
\subsection{Video Results}
Video results please refer to the supplementary materials.

\subsection{Results of Tree Search Visual Choice}
\begin{figure}[t]
    \centering
    \includegraphics[width=\linewidth]{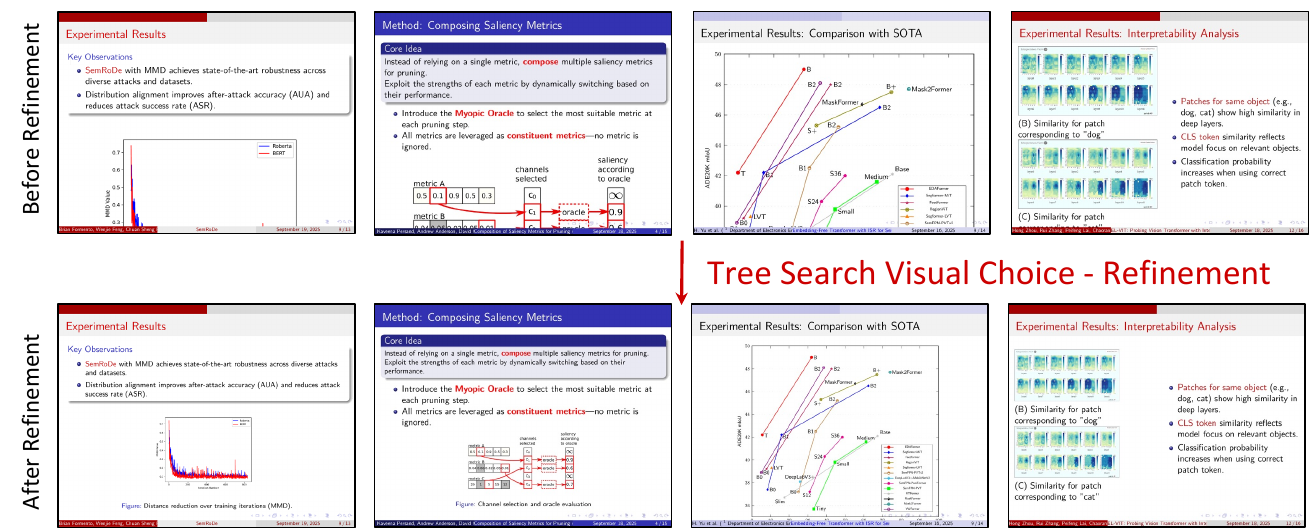}
    \caption{\textbf{Slide Visualization of Tree Search Visual Choice.} The first row shows slide results before layout refinement, while the second row shows their corresponding slides after refinement.}
    \label{fig:tree_vis}
\end{figure}

Figure~\ref{fig:tree_vis} illustrates the slides before and after applying tree search visual choice refinement. The refinement \textbf{resolves the overfull issues} and substantially improves slide quality, indicating that this module \textbf{plays a crucial role in layout adjustment}.

\section{Prompts}
% 颜色（可按需调整）
\definecolor{CardHeader}{RGB}{124,146,84}  % 橄榄绿标题栏
\definecolor{CardHeaderDark}{RGB}{100,120,65}
\definecolor{CardFrame}{RGB}{205,210,190}  % 浅绿边框

% 小圆点
\newcommand{\headerdot}{\tikz[baseline=-0.6ex]\node[circle,fill=CardHeaderDark,inner sep=1.5pt]{};}

% 卡片环境
\newtcolorbox{promptbox}[1]{%
  enhanced, breakable,
  colback=white, colframe=CardFrame,
  boxrule=0.5pt, arc=2mm,
  left=10pt,right=10pt,top=8pt,bottom=10pt,
  before skip=8pt, after skip=10pt,
  drop shadow=black!15,
  attach boxed title to top left={xshift=1mm,yshift*=-2mm},
  boxed title style={
    colback=CardHeader, colframe=CardHeader, coltext=white,
    boxrule=0pt, left=8pt,right=8pt,top=3pt,bottom=3pt, arc=2mm
  },
  title={\headerdot\ \textbf{Prompt:}~#1},
}

\begin{promptbox}{Slide Generation}
\textbf{System Prompt:} Please generate a complete English PPT introduction based on the following TeX source text content, using LaTeX Beamer. The specific requirements are as follows.

\medskip
\textbf{Content structure:}
\begin{itemize}[leftmargin=1.2em,itemsep=2pt,topsep=2pt]
  \item The PPT should contain the following chapters (arranged in order), and each chapter must have a clear title and content:
  \item Motivation (research background and problem statement)
  \item Related work (current status and challenges in the field)
  \item Method (core technical framework) [The content of the method needs to be introduced in detail, and each part of the method should be introduced on a separate page]
  \item Innovation (differentiation from existing work)
  \item Experimental method (experimental design and process)
  \item Experimental setting (dataset, parameters, environment, etc.)
  \item Experimental results (main experimental results and comparative analysis)
  \item Ablation experiment (validation of the role of key modules)
  \item Deficiencies (limitations of current methods)
  \item Future research (improvement direction or potential application)
  \item End slide (Thank you)
\end{itemize}

\textbf{Format requirements:}
\begin{itemize}[leftmargin=1.2em,itemsep=2pt,topsep=2pt]
\item Use Beamer's theme suitable for academic presentations, with simple color matching.
\item The content of each page should be concise, avoid long paragraphs, and use itemize or block environment to present points.
The title page contains the paper title, author, institution, and date.
\item Key terms or mathematical symbols are highlighted with \text{\\alert\{\}}.
\end{itemize}

\textbf{Image and table processing:}
\begin{itemize}[leftmargin=1.2em,itemsep=2pt,topsep=2pt]
\item All image paths are given, and relative paths are used when citing, the picture names must "be consistent with the name in tex file".
\item Images should automatically adapt to width, and add titles and labels
\item Experimental result tables should be extracted from the source text, formatted using tabular or booktabs environments, and marked with reference sources ( "as shown in table").
\end{itemize}

\textbf{Code generation requirements:}
\begin{itemize}[leftmargin=1.2em,itemsep=2pt,topsep=2pt]
\item The generated LaTeX code must be complete and can be compiled directly (including necessary structures).
\item Mark the source text location corresponding to each section in the code comments (for example, % corresponds to the source text Section 3.2).
\item If there are mathematical formulas in the source text, they must be retained and correctly converted to LaTeX syntax (such as $y=f(x)$).
\end{itemize}

\textbf{Other instructions:}
\begin{itemize}[leftmargin=1.2em,itemsep=2pt,topsep=2pt]

\item Image content should be read from the tex file, and the source name should be used directly without arbitrary modification. Image references should use real image names and should not be forged;
\item Table content should first extract real data from the source document.
\item All content should be in English.
\item If the source text is long, it is allowed to summarize the content, but the core methods, experimental data and conclusions must be retained.
\item To enhance readability, a transition page can be added (for example, "This section will introduce the experimental part").
\item Perfer more images than heavy text. **The number of slides should be around 10.** 
\item **\& in title is not allowed which will cause error "Misplaced alignment tab character \&\"**
**Pay attention to this "error: !File ended while scanning use of \text{\\frame}\"**
\item Only output latex code which should be ready to compile using tectonic(simple verson of TeX Live). Before output check if the code is grammatically correct.
\end{itemize}
\end{promptbox}

\begin{promptbox}{Error Correction}
\textbf{System Prompt:} You are given a LaTeX Beamer code for the slides of a research paper and its error information. Correct these errors \emph{without changing} the slide content (text, figures, layout).

\medskip
\textbf{Instructions:}
\begin{itemize}
  \item Apply the minimal edits required to make the file compile: add missing packages, close/open environments, balance braces, escape special characters, fix math delimiters, resolve duplicate labels, and correct obvious path or option typos.
  \item Do \emph{not} paraphrase or delete text; do \emph{not} change figure/table content, captions, labels, or layout semantics.
  \item Keep all image/table file names and relative paths as given; do not invent or rename assets.
  \item Preserve the original Beamer theme, colors, and structure.
  \item Ensure the final output compiles with \textbf{Tectonic}; close all environments and avoid undefined commands.
\end{itemize}

\medskip
\textbf{Output (strict):} Output \emph{only} the corrected LaTeX source, beginning with \texttt{beamer} and ending with \texttt{document}; no extra commentary.
\end{promptbox}

% 上面一行的说明文字（如截图左上角的粗体句子）
\begin{promptbox}{MSTS Judge}
\textbf{System Prompt:} You are a slide layout judge. You see four slides A–D in a 2×2 grid:
A (top-left), B (top-right), C (bottom-left), D (bottom-right).

\medskip
\textbf{Definitions}
\begin{itemize}[leftmargin=1.2em,itemsep=2pt,topsep=2pt]
  \item \textbf{Overfull:} any part of the figure or its caption is clipped, outside the frame, or overlapped/hidden.
  \item \textbf{Coverage:} among non-overfull options, larger visible content with less empty background is better.
  \item \textbf{Risk:} risk of overfull decreases from A → D (A largest, D smallest).
  \item \textbf{Coverage trend:} coverage decreases from A → D.
\end{itemize}

\textbf{Rules (judge only the given images)}
\begin{enumerate}[leftmargin=1.2em,itemsep=2pt,topsep=2pt]
  \item Disqualify any option with overfull (caption must be fully visible).
  \item From the remaining, pick the one with the greatest coverage.
  \item Practical method: scan \textbf{A → B → C → D}; choose the \emph{first} slide in that order that is not overfull.
\end{enumerate}

\medskip
\textbf{Output only (strict; do \emph{not} output \texttt{```json}):}
\begin{verbatim}
{
"reason": "concise comparison",
"choice": "A" | "B" | "C" | "D"
}
\end{verbatim}
\end{promptbox}

\begin{promptbox}{Slide Script with Cursor Positions}
\textbf{System Prompt:} You are an academic researcher presenting your own work at a research conference. You are provided with a sequence of adjacent slides.

\medskip
\textbf{Instructions:}
\begin{itemize}[leftmargin=1.2em,itemsep=2pt,topsep=2pt]
  \item For each slide, write a smooth, engaging, and coherent first-person presentation script.
  \item Clearly explain the \emph{current} slide with academic clarity, brevity, and completeness; use a professional, formal tone and avoid content unrelated to the paper.
  \item Each sentence must include \emph{exactly one} cursor position description drawn from the \emph{current slide} and listed in order, using the format \texttt{script\;|\;cursor description}. If no cursor is needed for a sentence, write \texttt{no}.
  \item Limit the total script for each slide to \textbf{50 words} or fewer.
  \item Separate slides using the delimiter \texttt{\#\#\#}.
\end{itemize}

\medskip
\textbf{Output Format (strict):}
\begin{verbatim}
sentence 1 | cursor description
sentence 2 | cursor description
...
###
sentence 1 | cursor description
...
\end{verbatim}
\end{promptbox}

\begin{promptbox}{Meta Similarity}
\textbf{System Prompt:} You are an evaluator. You will be given two presentation videos of the same talk: (1) a human-presented version and (2) an AI-generated version. Evaluate \emph{only} the slides and subtitles; ignore the presenter’s face, voice quality, background music, camera motion, and any non-slide visuals.

\medskip
\textbf{Inputs You May Receive}
\begin{itemize}[leftmargin=1.2em,itemsep=2pt,topsep=2pt]
  \item Human video (and optionally its slide images and subtitles/transcript)
  \item AI video (and optionally its slide images and subtitles/transcript)
\end{itemize}

\textbf{Evaluation Scope (focus strictly on slides + subtitles)}
\begin{enumerate}[leftmargin=1.2em,itemsep=2pt,topsep=2pt]
  \item \textbf{Slide Content Matching:} Do AI slides convey the same key points and comparable layout/visual elements (titles, bullets, diagrams, tables, axes annotations) as the human version?
  \item \textbf{Slide Sequence Alignment:} Is slide order consistent? Any sections missing, added, or rearranged?
  \item \textbf{Subtitle Wording Similarity:} Do AI subtitles reflect similar phrasing/terminology and information as the human speech/subtitles? Focus on semantic equivalence; minor style/spelling differences do not matter.
  \item \textbf{Slide–Subtitle Synchronization:} Within the AI video, does narration/subtitle content match the on-screen slide at the same time? Does this broadly align with the human presenter’s per-slide content?
\end{enumerate}

\textbf{Evidence-Only Rules}
\begin{itemize}[leftmargin=1.2em,itemsep=2pt,topsep=2pt]
  \item Base the judgment solely on the provided materials (videos, slides, subtitles). Do \emph{not} use outside knowledge.
  \item If some inputs are missing (\textit{e.g.}, no subtitles), judge from what is available and briefly note the missing piece in the Reasons.
\end{itemize}

\textbf{Relaxed Scoring Rubric (0–5)}
\begin{itemize}[leftmargin=1.2em,itemsep=2pt,topsep=2pt]
  \item \textbf{5} — Nearly identical: slides and subtitles closely match the human version in content, layout, sequence, and timing; wording is near-paraphrase.
  \item \textbf{4} — Highly similar: only minor layout/phrasing differences; content, order, and alignment clearly match.
  \item \textbf{3} — Moderate differences yet same core content: several layout/wording/sequence deviations but main sections and key points are preserved. (Leniency: borderline cases between 2 and 3 \emph{round up} to 3.)
  \item \textbf{2} — Partial overlap: substantial omissions/rearrangements or subtitle drift; multiple slide mismatches or sync issues.
  \item \textbf{1} — Minimal overlap: only a few matching fragments; most slides/subtitles diverge.
  \item \textbf{0} — No meaningful match: AI slides/subtitles do not correspond to the human version.
\end{itemize}
\noindent\emph{Lenient mapping: if borderline between adjacent levels, choose the higher score. If computing subscores, average and \textbf{round up} to the nearest integer in [0,5].}

\medskip
\textbf{Output Format (STRICT; exactly one line)}
\begin{verbatim}
Content Similarity: X/5; Reasons
\end{verbatim}
Where \texttt{X} is an integer 0–5 from the rubric, and \texttt{Reasons} is 1–3 short sentences referencing content, sequence, wording, and synchronization as relevant.
\end{promptbox}

\begin{promptbox}{PresentArena}
\textbf{System Prompt:} You are an expert in evaluating academic presentation videos. You are given two videos (Video A and Video B) on the same research topic. Evaluate each video independently and then decide which is better, or if they are basically the same (preferred when not confident).

\medskip
\textbf{Evaluation Criteria}
\begin{itemize}[leftmargin=1.2em,itemsep=2pt,topsep=2pt]
  \item \textbf{Content Clarity}: Are key ideas and findings clearly explained?
  \item \textbf{Speaker Delivery}: Is the speaker confident, fluent, and engaging?
  \item \textbf{Visual Aids}: Are slides/visuals clear, helpful, and well-integrated?
  \item \textbf{Structure \& Pacing}: Is the talk logically organized and appropriately paced?
  \item \textbf{Audience Engagement}: Does the speaker maintain interest and attention?
\end{itemize}

\textbf{Steps}
\begin{enumerate}[leftmargin=1.2em,itemsep=2pt,topsep=2pt]
  \item \textbf{Step 1:} Write a short (1–2 sentence) evaluation of \textbf{Video A} based on the criteria.
  \item \textbf{Step 2:} Write a short (1–2 sentence) evaluation of \textbf{Video B} based on the criteria.
  \item \textbf{Step 3:} Decide which video is better, or if they are basically the same (prefer ``Same'' if not confident).
\end{enumerate}

\medskip
\textbf{Output Format (Strict; only these three blocks):}
\begin{verbatim}
Step 1:
[1–2 sentences evaluating Video A]

Step 2:
[1–2 sentences evaluating Video B]

Step 3:
Final Judgment:
[A] | [B] | [Same]

Reason: [One concise sentence justifying the judgment based on Steps1-2.]
\end{verbatim}
\end{promptbox}

\begin{promptbox}{PresentationQuiz}
\textbf{System Prompt:} You are an answering agent. You will be provided with:
1) a presentation video of a paper, and 2) a JSON object called \texttt{"questions"} containing multiple questions, each with four options (A–D).
Analyze the video thoroughly and answer each question \emph{solely} based on the video content (no external knowledge). Do not reference timesteps that exceed the video length.

\medskip
\textbf{Instructions:}
\begin{itemize}
  \item For each question, if the video provides sufficient evidence for a specific option (A, B, C, or D), choose that option.
  \item Include a brief reference to where in the video the evidence appears (\textit{e.g.}, ``Top-left text'', ``Event date section'').
  \item Rely only on the video; do not use outside context.
  \item Provide an answer entry for \emph{all} questions present in \texttt{"questions"}.
\end{itemize}

\medskip
\textbf{Template (steps to follow):}
\begin{enumerate}
  \item Study the presentation video together with \texttt{"questions"}.
  \item For each question, determine whether the video clearly supports one of the four options; if so, pick that answer.
  \item Provide a brief reference indicating where in the video you found the evidence.
  \item Format the final output strictly as a JSON object with the following pattern (and no extra keys or explanations).
\end{enumerate}

\textbf{Output Format (strict):}
\begin{verbatim}
{
  "Question 1": { "answer": "X", "reference": "some reference" },
  "Question 2": { "answer": "X", "reference": "some reference" },
  ...
}
\end{verbatim}

\textbf{questions payload:}
\begin{verbatim}
{{questions}}
\end{verbatim}
\end{promptbox}

%%%%%%%%%%%%%%%%%%%%%%%%%%%%%%%%%%%%%%%%%%%%%%%%%%%%%%%%%%%%

% \appendix

% \section{Technical Appendices and Supplementary Material}
% Technical appendices with additional results, figures, graphs and proofs may be submitted with the paper submission before the full submission deadline (see above), or as a separate PDF in the ZIP file below before the supplementary material deadline. There is no page limit for the technical appendices.

%%%%%%%%%%%%%%%%%%%%%%%%%%%%%%%%%%%%%%%%%%%%%%%%%%%%%%%%%%%%
\end{document}